\documentclass{article}

\PassOptionsToPackage{numbers, compress}{natbib}

\usepackage[final]{neurips_2020}

\usepackage[utf8]{inputenc} 
\usepackage[T1]{fontenc}    
\usepackage{hyperref}       
\usepackage{url}            
\usepackage{booktabs}       
\usepackage{amsfonts}       
\usepackage{nicefrac}       
\usepackage{microtype}      

\usepackage{subfiles}
\usepackage{lifetime}

\usepackage{graphicx}
\usepackage{subfigure}
\usepackage{wrapfig}
\title{A Unified View of Label Shift Estimation}

\author{Saurabh Garg, Yifan Wu, Sivaraman Balakrishnan, Zachary C. Lipton \\
Machine Learning Department, \\
Department of Statistics and Data Science, \\
Carnegie Mellon University\\
\small{\texttt{\{\href{mailto:sgarg2@andrew.cmu.edu}{sgarg2},\href{mailto:yw4@andrew.cmu.edu}{yw4},\href{mailto:sbalakri@andrew.cmu.edu}{sbalakri},\href{mailto:zlipton@andrew.cmu.edu}{zlipton}\}@andrew.cmu.edu
}}}

\begin{document}

\maketitle

\begin{abstract}
Under label shift, the label distribution $p(y)$ 
might change but the class-conditional distributions $p(x|y)$ do not.
There are two dominant approaches for estimating the label marginal. 
BBSE, a moment-matching approach based on confusion matrices, 
is provably consistent and provides interpretable error bounds. 
However, a maximum likelihood estimation approach, 
which we call MLLS, dominates empirically.  
In this paper, we present a unified view of the two methods and 
the first theoretical characterization of MLLS.
Our contributions include 
(i) consistency conditions for MLLS, 
which include calibration of the classifier and a 
confusion matrix invertibility condition that BBSE also requires; 
(ii) a unified framework, casting 
BBSE as roughly equivalent to MLLS 
for a particular choice of calibration method; 
and (iii) a decomposition of MLLS's finite-sample error into 
terms reflecting 
miscalibration and estimation error. 
Our analysis attributes BBSE's statistical inefficiency 
to a loss of information due to coarse calibration. 
Experiments on synthetic data, MNIST, and CIFAR10 
support our findings.

\end{abstract}


\vspace{-10pt}
\section{Introduction}
\vspace{-5pt}
Supervised algorithms are typically developed and evaluated 
assuming independent and identically distributed (iid) data.  
However, the real world seldom abides,
presenting domain adaptation problems 
in which the \textit{source distribution} $\ProbS$,
from which we sample labeled training examples, 
differs from the \textit{target distribution} $\ProbT$,
from which we only observe unlabeled data.
Absent assumptions on the nature of shift, 
the problem is 
underspecified.
Multiple assumptions may be compatible 
with the same observations while implying
different courses of action.
Fortunately, some assumptions can render 
shift detection, estimation, and on-the-fly updates 
to our classifiers possible.

This paper focuses on \emph{label shift}
\citep{storkey2009training, saerens2002adjusting, lipton2018detecting},
which aligns with 
the \emph{anticausal} setting 
in which the labels $y$ cause the features $\x$ \citep{scholkopf2012causal}.
Label shift arises in diagnostic problems
because diseases cause symptoms.
In this setting, an intervention on $p(y)$ induces the shift,
but the process generating $x$ given $y$ is fixed ($\ps(\x|y) = \pt(\x|y)$).
Under label shift, the optimal predictor 
may change,
e.g., the probability that a patient suffers from a disease
given their symptoms can increase under a pandemic.
Contrast label shift with the better-known 
\emph{covariate shift} assumption, 
which aligns with the assumption that $\x$ causes $y$,
yielding the reverse implication that $\ps(y|\x) = \pt(y|\x)$.

Under label shift, our first task 
is to estimate the ratios
$w(y) = \pt(y)/\ps(y)$ for all labels $y$.
%
%
Two dominant approaches leverage off-the-shelf 
classifiers to estimate $\w$: (i) \emph{Black Box Shift Estimation} (BBSE) ~\citep{lipton2018detecting} and a variant called \emph{Regularized Learning under Label Shift} (RLLS)~\citep{azizzadenesheli2019regularized}: 
moment-matching based estimators that leverage 
(possibly biased, uncalibrated, or inaccurate) 
predictions to estimate the shift; 
and (ii) Maximum Likelihood Label Shift (MLLS) \citep{saerens2002adjusting}: 
an Expectation Maximization (EM) algorithm that 
assumes access to a classifier that outputs
the true source distribution conditional probabilities~$\ps(y|\x)$.

%
%
%
%
%
%
%
%
%
Given a predictor $\smash{\widehat{\f}}$
with an invertible confusion matrix,
BBSE and RLLS have known consistency results
and finite-sample guarantees 
\citep{lipton2018detecting, azizzadenesheli2019regularized}.
However, 
%
%
MLLS, in combination with
a calibration heuristic called Bias-Corrected Temperature Scaling (BCTS),
outperforms them empirically \citep{alexandari2019adapting}.



In this paper, we theoretically characterize MLLS, 
establishing conditions for consistency 
and bounding its finite-sample error.
To start, we observe that given the true label conditional $\ps(y|\x)$, 
MLLS is simply a concave Maximum Likelihood Estimation (MLE) problem 
and standard results apply.
However, because we never know $\ps(y|\x)$ exactly, 
MLLS is always applied with an estimated model $\smash{\widehat{\f}}$ 
and thus the procedure consists of MLE under model misspecification.


First, we prove that (i) \emph{\calib calibration} (Definition~\ref{def:calib}) 
and (ii) an invertible confusion matrix (as required by BBSE) 
are \emph{sufficient conditions} to ensure MLLS's consistency (Proposition~\ref{prop:equiv-cond}, Theorems~\ref{theorem:mlls} and~\ref{theorem:convergence}).
We also show that calibration can sometimes be \emph{necessary}
for consistency (Example~1 in Section~\ref{subsec:MLLS}). 
Recall that neural network classifiers tend to be uncalibrated 
absent post-hoc adjustments \cite{guo2017calibration}.
%
Second, we observe that confusion matrices
can be instruments for calibrating a classifier. 
Applying MLLS with this technique,
BBSE and MLLS are distinguished only
by their objective functions. 
Through extensive experiments, 
we show that they
perform similarly, 
concluding that MLLS's superior performance
(when applied with more granular calibration techniques)
is not due to its objective but rather 
to the information lost by BBSE
via confusion matrix calibration. 
Third, we analyze the finite-sample error 
of the MLLS estimator by decomposing its error into terms 
reflecting the miscalibration error and finite-sample error (Theorem~\ref{theorem:main-error-bound}). 
Depending on the calibration method, 
the miscalibration error can further be divided into two terms: 
finite sample error due to re-calibration on a validation set 
and the minimum achievable calibration error with that technique. 

We validate our results on synthetic data,
MNIST, and CIFAR-10.
Empirical results show that MLLS can have
$2$--$10\times$ lower Mean Squared estimation Error (MSE)
depending on the magnitude of the shift. 
Our experiments relate MLLS's MSE 
to the granularity of the calibration. 

In summary, we contribute the following:
(i) Sufficient conditions for MLLS's consistency;
(ii) Unification of MLLS and BBSE methods under a common framework, 
with BBSE corresponding to a particular choice of calibration method;
(iii) Finite-sample error bounds for MLLS;  
(iv) Experiments on synthetic and image recognition datasets 
that support our theoretical arguments. 

\vspace{-8pt}
\section{Problem Setup}
\vspace{-5pt}
\label{sec:pre}

Let $\inpt$ be the input space and 
$\out = \{1,2,\ldots, k\}$ the output space. 
Let $\ProbS, \ProbT : \inpt\times\out \to [0,1]$
be the source and target distributions
and let $\ps$ and $\pt$ denote the corresponding 
probability density (or mass) functions.
We use $\mathbb{E}_s$ and $\mathbb{E}_t$ to denote expectations 
over the source and target distributions. 
%
%
In unsupervised domain adaptation, 
we possess labeled source data 
$\{(\x_1,y_1), (\x_2, y_2), \ldots , (\x_n, y_n)\}$ 
and unlabeled target data $\{\x_{n+1}, \x_{n+2}, \ldots, \x_{n+m}\}$. 
%
We also assume access to a black-box predictor 
$\smash{\widehat \f: \inpt \mapsto \Delta^{k-1}}$, e.g., 
a model trained to approximate 
the true probability function $f^*$, 
where $\smash{f^*(x) \defeq \ps(\cdot |\x)}$. 
Here and in the rest of the paper, 
we use $\Delta^{k-1}$ to denote the standard $k$-dimensional probability simplex. 
For a vector $\vv$, we use $\vv_y$ 
to access the element at index $y$. 

%
%


Absent assumptions relating the source and target distributions,
domain adaptation is underspecified \citep{ben2010impossibility}.
We work with the \emph{label shift} assumption, 
i.e., $\ps(\x|y) = \pt(\x|y)$,
focusing on multiclass classification.
Moreover, we assume non-zero support 
for all labels in the source distribution: 
for all $y \in \out$, $\ps(y) \ge c > 0$ \citep{lipton2018detecting,azizzadenesheli2019regularized}. 
Under label shift, three common goals are 
(i) detection---determining whether distribution shift has occurred;
(ii) quantification---estimating
the target label distribution; 
and (iii) correction---producing 
a predictor that minimizes error on the target distribution
\citep{lipton2018detecting}.
 
%
%
%
This paper focuses on goal (ii),
estimating importance weights $w(y) = \pt(y)/\ps(y)$ 
for all $y \in \out$.
%
Given $\w$, we can update our classifiers on the fly,
either by retraining in an importance-weighted ERM framework \citep{shimodaira2000improving, gretton2009covariate, lipton2018detecting, azizzadenesheli2019regularized}---a practice 
that may be problematic for
overparameterized neural networks \citep{byrd2019effect},
or by applying an analytic correction
\citep{alexandari2019adapting, saerens2002adjusting}.   
%
%
%
%
%
Within the ERM framework, the generalization result from 
\citet{azizzadenesheli2019regularized} (Theorem 1)
depends only on the error of the estimated weights, 
and hence any method that improves weight estimates
tightens this bound. 

%
%
%
%
%
%
%

%
%
%

There are multiple definitions of calibration in the multiclass setting.
\citet{guo2017calibration} study
the calibration of the arg-max prediction,
while \citet{kumar2019verified} study a notion of 
per-label calibration.  
We use
\calib calibration~\citep{vaicenavicius2019evaluating}
and the expected \calib calibration error on the source data defined as follows:  
\begin{definition}[Canonical calibration]~\label{def:calib}
 A prediction model $\f: \mathcal{X} \mapsto \Delta^{k-1}$ is canonically calibrated  
 on the source domain 
 if for all $\x \in \inpt$ and $j \in \out$, 
 $\mathrm{P}_s ( y = j | \f(x) ) = \f_j(x) \,.$
\end{definition}
\begin{definition}[Expected \calib calibration error]~\label{def:caliberrr}
For a predictor $\f$, the expected squared \calib calibration error on the source domain is 
 $\smash{\mathcal{E}^2(f) =   \mathbb{E}_s \norm{\f - \fc}^2}$,
 where $\smash{\fc = \mathrm{P}_s(y = \cdot |f(\x))}$.
\end{definition}

Calibration methods typically work 
either by calibrating the model during training 
or by calibrating a trained classifier on held-out data, post-hoc.
We refer the interested reader to~\citet{kumar2019verified} and~\citet{guo2017calibration} for detailed studies on calibration. 
We focus on the latter category of methods. Our experiments follow \citet{alexandari2019adapting},
who leverage BCTS \footnote{Motivated by the strong empirical results in \citet{alexandari2019adapting}, 
we use BCTS in our experiments as a surrogate to canonical calibration.}
to calibrate their models. BCTS extends temperature scaling~\citep{guo2017calibration} by incorporating 
per-class bias terms.

\vspace{-8pt}
\section{Prior Work}
\vspace{-5pt}
\label{sec:prior} 
Two families of solutions have been explored that leverage a blackbox predictor:
BBSE \citep{lipton2018detecting}, 
a moment matching method,
uses the predictor $\smash{ \widehat \f}$ to compute a confusion matrix 
$\smash{ C_{\widehat{\f}} \defeq \ps(\widehat{y}, y) \in \Real^{k\times k}}$ on the source data. 
Depending on how $\smash{\widehat{y}}$ is defined, 
there are two types of confusion matrix for a predictor $\smash{\widehat{\f}}$: 
(i) the \emph{hard confusion matrix} $\smash{\widehat{y}=\argmax \widehat{\f}(\x)}$; and (ii) the \emph{soft confusion matrix}, where $\smash{\widehat{y}}$ is defined 
as a random prediction that follows 
the discrete distribution $\smash{\widehat{\f}(\x)}$ over $\smash{\out}$.
Both soft and hard confusion matrix 
can be estimated from labeled source data samples.
The estimate $\smash{\widehat \w}$ is computed as 
$\smash{\widehat \w := \widehat C_{\widehat \f}^{-1} \widehat \mu}$, 
where $\smash{\widehat C_{\widehat{\f}}}$ is the estimate of confusion matrix and $\smash{\widehat\mu}$ is an 
estimate of $\smash{\pt(\widehat y)}$,
%
computed by applying the predictor $\smash{\widehat \f}$ to the target data.
%
In a related vein, RLLS~\citep{azizzadenesheli2019regularized} 
incorporates an additional regularization term of the form $\norm{\w - 1}$ 
and solves a constrained optimization problem 
to estimate the shift ratios $\w$. 

%
%
MLLS estimates $w$ as if performing maximum likelihood estimation,
but substitutes the predictor outputs for the true probabilities $\ps(y|\x)$.
\citet{saerens2002adjusting}, who introduce this procedure,
describe it as an application of EM.
However, as observed in \citep{du2014semi, alexandari2019adapting}, 
the likelihood objective is concave, 
and thus a variety of 
optimization algorithms may be applied to recover the MLLS estimate.  
\citet{alexandari2019adapting} also showed that MLLS underperforms BBSE when applied naively, a phenomenon that we shed 
more light on in this paper.




\vspace{-8pt}
\section{A Unified View of Label Shift Estimation with Black Box Predictors} \label{sec:unify}
\vspace{-5pt}
\newcommand{\ltt}{\mathcal{Z}}
\newcommand{\W}{\mathcal{W}}
\newcommand{\z}{z}
%
We now present a unified view that subsumes MLLS and BBSE
%
%
%
and demonstrate how each is instantiated under this framework. 
We also establish identifiability and consistency conditions for MLLS, 
deferring a treatment of finite-sample issues to Section \ref{sec:finite}. 
For convenience, throughout Sections 3 and 4, 
we use the term \emph{calibration} exclusively 
to refer to canonical calibration 
(Definition~\ref{def:calib}) on the source data. 
We relegate all technical proofs to Appendix~\ref{sec:AppendixA}. 

\vspace{-5pt}
\subsection{A Unified Distribution Matching View}
\vspace{-5pt}

To start, we introduce a \emph{generalized} distribution
matching approach for estimating $\w$. 
Under label shift, for any (possibly randomized) mapping 
from $\inpt$ to $\ltt$, we have that
$p_s(\z|y)=p_t(\z|y)$ since,
$\smash{p_s(\z|y)=p_t(\z|y)=\int_{\mathcal{X}} \, p(\z|\x)p(\x|y)dx.}$
%
%
Throughout the paper, we use the notation $p(\z|y)$
to represent either $p_s(\z|y)$ or $p_t(\z|y)$ (which are identical). 
We now define a family of distributions over 
$\ltt$ parameterized by $\w\in \W$ as
\begin{align}
\label{eqn:pwdef}
p_{\w}(\z) = \sum\nolimits_{y=1}^k p(\z|y)p_s(y)\w_y = \sum\nolimits_{y=1}^k p_s(\z, y)\w_y, 
\end{align}
where $\smash{\W = \{ \w \; |\; \forall y \,,  w_y \ge 0 \text{ and } \sum_{y=1}^{k} w_y \ps(y) = 1 \}}$.     
When 
$\w=\w^*$, we have that $p_{\w}(\z) = p_t(\z)$. For fixed $p(\z|\x)$, $p_t(\z)$ and $p_s(\z, y)$ are known because $p_t(\x)$ and $p_s(\x, y)$ are known. 
So one potential strategy to estimate $\w^*$ is 
to find a weight vector $\w$ such that 
\begin{align*}
\sum\nolimits_{y=1}^k p_s(\z, y)\w_y = p_t(\z) ~~~~ \forall \z\in \ltt \,. \addeq\label{eq:distr-match}
\end{align*}
At least one such weight vector $\w$ must exist as $\w^*$ satisfies \eqref{eq:distr-match}. 
We now characterize
conditions under which the weight vector $\w$ satisfying \eqref{eq:distr-match} is unique:

\begin{restatable}[Identifiability]{relemma}{identifiability}
\label{lemma:identi-1}
If the set of distributions $\{p(\z|y)\,:\, y=1,...,k\}$ are linearly independent,
then for any $\w$ that satisfies \eqref{eq:distr-match}, we must have $\w=\w^*$. 
This condition is also necessary in general: 
if the linear independence does not hold 
then there exists a problem instance where 
we have $\w, \w^* \in \W$ satisfying \eqref{eq:distr-match} while $\w \ne \w^*$. 
\end{restatable}

Lemma~\ref{lemma:identi-1} follows from the fact 
that \eqref{eq:distr-match} is a linear system 
with at least one solution $\w^*$. This solution is 
unique when $p_s(\z, y)$ is of rank $k$.
The linear independence condition in Lemma~\ref{lemma:identi-1}, in general, is sufficient for identifiability of discrete $\ltt$. 
%
%
However, for continuous $\ltt$, the linear dependence condition has the undesirable property of being sensitive to changes on sets of measure zero.
By changing a collection of linearly dependent distributions on a set of measure zero, we can make them linearly independent.
As a consequence, we impose a \emph{stronger} notion of identifiability i.e., the set of distributions $\{p(\z|y)\,:\, y=1,...,k\}$ are such that there does not exist $\vv \ne 0$ for which 
    $\smash{\int_{\ltt} \lvert{\sum_y p(\z|y) v_y}\rvert d\z = 0.}$ 
We refer 
this condition as \emph{strict linear independence}. 

In generalized distribution matching, one can set $p(\z|\x)$ to be the Dirac delta function 
at $\delta_{\x}$\footnote{For simplicity we will use $\z=\x$ 
to denote that $p(\z|\x)=\delta_{\x}$.}
such that $\ltt$ is the same space as $\inpt$, 
which leads to solving \eqref{eq:distr-match} with $\z$ replaced by $\x$. 
In practice where $\inpt$ is high-dimensional and/or continuous,
approximating the solution to \eqref{eq:distr-match} from finite samples
can be hard when choosing $\z=\x$. 
Our motivation for generalizing
distribution matching from $\inpt$ to 
$\ltt$ is that the solution to \eqref{eq:distr-match} 
can be better approximated using finite samples 
when $\ltt$ is chosen carefully.
Under this framework, the design of a label shift 
estimation algorithm can be decomposed into two parts: 
(i) the choice of $p(\z|\x)$ and
(ii) how to approximate the solution to \eqref{eq:distr-match}. 
Later on, we consider how these design choices 
may affect label shift estimation procedures in practice.

\vspace{-7pt}
\subsection{The Confusion Matrix Approach} 
\vspace{-5pt}

If $\ltt$ is a discrete space, one can first estimate 
$p_s(\z, y) \in \Real^{|\ltt|\times k}$ and 
$p_t(\z) \in \Real$, and then subsequently attempt 
to solve \eqref{eq:distr-match}.
Confusion matrix approaches use $\ltt = \out$, 
and construct $p(\z|\x)$ using a black box predictor $\widehat{\f}$. 
There are two common choices to construct the confusion matrix:
%
(i) The soft confusion matrix approach:
We set $p(\z|\x) := \widehat{\f}(\x) \in \Delta^{k-1}$. 
We then define a random variable $\widehat{y} \sim \widehat{\f}(\x)$ for each $\x$. 
Then we construct $p_s(\z, y) = p_s(\widehat{y}, y)$ and $p_t(\z)=p_t(\widehat{y})$. 
(ii) The hard confusion matrix approach: Here we set $p(\z|\x)=\delta_{\argmax\widehat{\f}(\x)}$. 
We then define a random variable $\widehat{y} =\argmax \widehat{\f}(\x)$ for each $\x$. 
Then again we have $p_s(\z, y) = p_s(\widehat{y}, y)$ and $p_t(\z)=p_t(\widehat{y})$. 

Since $p_s(\z, y)$ is a square matrix, 
the identifiability condition becomes 
the invertibility of the confusion matrix. 
Given an estimated confusion matrix, one can find $\w$ by inverting the confusion matrix (BBSE) or minimizing
some distance 
between the vectors on the two sides of \eqref{eq:distr-match}. 


\vspace{-7pt}
\subsection{Maximum Likelihood Label Shift Estimation} \label{subsec:MLLS}
\vspace{-5pt}

When $\ltt$ is a continuous space, 
the set of equations in \eqref{eq:distr-match} indexed by $\ltt$ is intractable. 
In this case, 
one possibility is to 
find a weight vector $\widetilde{\w}$
by minimizing the KL-divergence
$\mathrm{KL}(p_t(\z), p_{\w}(\z)) = \EE{t}{\log p_t(\z) / p_{\w}(\z)}$, for 
$p_{\w}$ defined in~\eqref{eqn:pwdef}.
This is equivalent to maximizing the population log-likelihood:
$\smash{\widetilde{w} := \argmax_{w \in \W} \EE{t}{ \log p_{\w}(\z)}}\,.$
One can further show that 
\allowdisplaybreaks
$\EE{t}{ \log p_{\w}(\z)}
= \mathbb{E}_{t}[\log \sum_{y=1}^k p_s(\z, y)\w_y]  = \mathbb{E}_{t}[ \log \sum_{y=1}^k p_s(y|\z) p_s(\z) \w_y]
= \mathbb{E}_{t}[ \log \sum_{y=1}^k p_s(y|\z)  \w_y] + \EE{t}{\log p_s(\z)}.$
Therefore we can equivalently define:
\begin{align*}
\widetilde{w} := \argmax_{w \in \W} \mathbb{E}_t \Big[ \log \sum\nolimits_{y=1}^k p_s(y|\z)  \w_y\Big] \,.\addeq\label{eq:mlls-z}
\end{align*}
This yields a straightforward convex optimization problem
whose objective is bounded from below
\citep{alexandari2019adapting, du2014semi}. 
Assuming access to labeled source data 
and unlabeled target data,
one can maximize 
the empirical counterpart of the objective in~\eqref{eq:mlls-z},
using either EM or an alternative iterative optimization scheme.
\citet{saerens2002adjusting} derived an EM algorithm to maximize the objective \eqref{eq:mlls-z} when $\z=\x$, 
assuming access to $p_s(y|\x)$. 
Absent knowledge of the ground truth $p_s(y|\x)$, 
we can plug in any approximate predictor $\f$ 
and optimize the following objective:
\begin{align*}
w_f := \argmax_{w \in \W} \LL(\w,\f) := \argmax_{w \in \W} \EE{t}{ \log \f(\x)^T  \w} \,.\addeq\label{eq:mlls-main}
\vspace{-5pt}
\end{align*}
%
%
%
In practice, $\f$ is fit 
from a finite 
sample drawn from $p_s(\x, y)$ 
and standard machine learning methods
often produce uncalibrated predictors.
While BBSE and RLLS are provably consistent 
whenever the predictor $\f$ yields an invertible confusion matrix, 
%
%
to our knowledge, no prior works have established sufficient conditions 
to guarantee  MLLS' consistency when $\f$ differs from $p_s(y|\x)$.

It is intuitive that for some values of $\f \neq p_s(y|\x)$, MLLS will yield inconsistent estimates. 
Supplying empirical evidence, \citet{alexandari2019adapting} show that MLLS performs poorly 
when $\f$ is a vanilla neural network predictor learned from data.
However, \citet{alexandari2019adapting} also show that
in combination with a particular post-hoc calibration technique,
MLLS achieves low error, significantly outperforming BBSE and RLLS. 
As the calibration error is not a distance metric between $\f$ and $p_s(y|\x)$ 
(zero calibration error does not indicate $\f=p_s(y|\x)$),
a calibrated predictor $\f$ may still be substantially different from $p_s(y|\x)$. 
Some natural questions then arise: 

\begin{enumerate}[noitemsep,topsep=0pt]
\setlength\itemsep{4pt}
\item \emph{Why does calibration improve MLLS so dramatically?}
\item \emph{Is calibration necessary or sufficient to ensure the consistency of MLLS?}
\item \emph{What accounts for the comparative efficiency of MLLS 
over BBSE?} (Addressed in Section~\ref{sec:finite})
\end{enumerate}



To address the first two questions, 
we make the following observations. 
Suppose we define $z$ (for each $x$) with distribution 
$p(\z| \x) := \delta_{\f(\x)}$, for some calibrated predictor $f$. 
Then, because $f$ is calibrated,
it holds that $p_s(y | \z) = f(x)$.
%
%
Note that in general,
the MLLS objective~\eqref{eq:mlls-main} can 
differ from~\eqref{eq:mlls-z}. 
However, when $p(\z| \x) := \delta_{\f(\x)}$, 
the two objectives are identical.
We can formalize this as follows:
\begin{restatable}[]{relemma}{calibobj}
If $\f$ is calibrated, then the two objectives~\eqref{eq:mlls-z} and~\eqref{eq:mlls-main}
are identical when $\ltt$ is chosen as $\Delta^{k-1}$ and \mbox{$p(\z|\x)$} is defined to be {$\delta_{\f(\x)}$}.
\label{lemma:calib}
\end{restatable}
Lemma~\ref{lemma:calib} follows from changing 
the variable of expectation in \eqref{eq:mlls-main} from $\x$ to $\f(\x)$ 
and applying $\f(\x)=p_s(y|\f(\x))$ (definition of calibration). 
It shows that MLLS with a calibrated predictor on the input space $\inpt$ 
is in fact equivalent to performing 
distribution matching in the space $\ltt$.
Building on this observation, we now state our 
population-level consistency theorem for MLLS:

\begin{restatable}[Population consistency of MLLS]{rethm}{consistency}
If a predictor $\f: \inpt \mapsto \Delta^{k-1}$ is calibrated and the distributions 
$\{p(\f(\x)|y)\,:\,y=1,\ldots,k\}$ are strictly linearly independent, 
then $\w^*$ is the unique maximizer of the MLLS objective~\eqref{eq:mlls-main}.
\label{theorem:mlls}
\end{restatable}

We now turn our attention to establishing consistency of the sample-based estimator.
Let \allowdisplaybreaks{$\x_1, \x_2, \ldots, \x_m \stackrel{iid}{\sim} \pt(x)$}. 
The finite sample objective for MLLS can be written as
\begin{align*}
\widehat{w}_f := \argmax_{w \in \W} \frac{1}{m} \sum\nolimits_{i=1}^m \log \f(\x_i)^T  \w := \argmax_{w \in \W} \LL_m(\w, \f) \,.\addeq\label{eq:mlls-sample}
\end{align*}
\begin{restatable}[Consistency of MLLS]{rethm}{convergence}
\label{theorem:convergence}
If $\f$ satisfies the conditions in Theorem~\ref{theorem:mlls}, then $\widehat{w}_f$ in~\eqref{eq:mlls-sample} converges to $\w^*$ almost surely. 
\end{restatable}

The main idea of the proof of Theorem~\ref{theorem:convergence} is to derive a metric entropy bound on the class of functions $\mathcal{G} = \left\{ (f^Tw)/(f^Tw + f^Tw^*) | w\in \mathcal{W} \right\}$ to prove Hellinger consistency (Theorem 4.6~\citep{geer2000empirical}). 
The consistency of MLLS 
relies on the linear independence of
the collection of distributions $\{p(\f(\x)|y)\,:\,y=1,\ldots,k\}$. 
The following result develops several alternative equivalent characterizations 
of this linear independence condition. 

\begin{restatable}[]{reprop}{equivalent} 
\label{prop:equiv-cond}
For a calibrated predictor $\f$, the following statements are equivalent:
\begin{enumerate}[itemsep=-.5ex,topsep=0pt,label=(\arabic*)]
\item $\{p(\f(\x)|y)\,:\,y=1,\ldots,k\}$ are strictly linearly independent.
\item $\EE{s}{\f(\x)\f(\x)^T}$ is invertible.
\item The soft confusion matrix of $\f$ is invertible.
\end{enumerate}
\end{restatable}

Proposition~\ref{prop:equiv-cond} shows that with a calibrated predictor,
the invertibility condition as required by BBSE (or RLLS)
is exactly the same as the linear independence condition 
required for MLLS's consistency.





 
Having provided sufficient conditions,  
we consider a binary classification example
to provide intuition for why we need calibration for consistency.
In this example, we relate the estimation error to the miscalibration error,
showing that calibration is not only sufficient but also necessary
to achieve zero estimation error for a certain class of predictors. 
%
%
%
%

\textbf{Example 1.} Consider a mixture of two Gaussians with 
\allowdisplaybreaks{$\ps(x|y=0) \defeq \N(\mu, 1)$ 
and $\ps(x|y=1) \defeq \N(-\mu, 1)$}. 
We suppose that the source mixing coefficients are both $\frac{1}{2}$, while
the target mixing coefficients are $\alpha (\ne \frac{1}{2}), 1-\alpha $. 
Assume a class of probabilistic threshold classifiers:
$f(x) = [1-c, c]$ for $x\ge 0$, otherwise $f(x) = [c, 1-c]$ with $c \in [0,1]$.
%
Then the population error of MLLS is given by 
\begin{align*}
4\abs{\frac{(1-2\alpha)(\ps(x\ge0|y=0) - c)}{1-2c}},
\end{align*}
which is zero only if $c = \ps(x\ge0|y=0)$ for a non-degenerate classifier.


The expression for estimation error arising from our example yields two key insights:
(i) an uncalibrated thresholded classifier has an estimation error 
proportional to the true shift in label distribution  i.e. $1-2\alpha$;
(ii) the error is also proportional to the 
canonical calibration error which is $\ps(x\ge0|y=0) - c$. 
%
While earlier in this section, we concluded 
that canonical calibration is sufficient for consistency,
the above example provides some intuition 
for why it might also be necessary. 
In Appendix~\ref{ref:marg_calib}, we show that marginal calibration~\citep{kumar2019verified,guo2017calibration,vaicenavicius2019evaluating}, a less restricted definition is insufficient to achieve consistency. 

\vspace{-5pt}
\subsection{MLLS with Confusion Matrix} \label{subsec:MLLS_CM}
\vspace{-3pt}

So far, we have shown that MLLS with any calibrated predictor 
can be viewed as distribution matching in a latent space. 
Now we discuss a method to construct a predictor $\f$ 
to perform MLLS given any $p(\z|\x)$, e.g.,
those induced by confusion matrix approaches. 
Recall, we already have the maximum log-likelihood objective.
It just remains to construct a calibrated predictor $\f$ from the confusion matrix. 

This is straightforward when $p(\z|\x)$ is deterministic, i.e.,
$p(\z|\x) = \delta_{g(\x)}$ for some function $g$: 
setting $\f(\x) = p_s(y|g(\x))$ makes the objectives \eqref{eq:mlls-z} and \eqref{eq:mlls-main} to be the same. 
Recall that for the hard confusion matrix,
the induced latent space is 
$p(\z|\x) = \delta_{\argmax \widehat{\f}(\x)}$.
So the corresponding predictor in MLLS is $\smash{\f(\x) = p_s(y|\widehat{y}_{\x})}$,
where $\smash{\widehat{y}_{\x} = \argmax \widehat{\f}(\x)}$.
Then we obtain the MLLS objective for the hard confusion matrix:
\vspace{-5pt}
\begin{align*}
\max_{w \in \W} \EE{t}{ \log \sum\nolimits_{y=1}^k p_s(y|\widehat{y}_{\x})  \w_y} \,.
\addeq\label{eq:cm-log} 
\end{align*}
The confusion matrix $\smash{\C_{\widehat{\f}}}$ and predictor $\smash{\widehat{\f}}$
directly give us $\smash{p_s(y|\widehat{y}_{\x})}$. Given an input $\x$, one can first get $\widehat{y}_{\x}$ from $\widehat{\f}$, 
then normalize the $\widehat{y}_{\x}$-th row of $\C_{\widehat{\f}}$ as $p_s(y|\widehat{y}_{\x})$.  We denote MLLS with hard confusion matrix calibration \eqref{eq:cm-log} by MLLS-CM. 

When $p_s(\z|\x)$ is stochastic, we need to extend \eqref{eq:mlls-main} to allow $\f$ to be a random predictor: 
$\f(\x)=p_s(y|\z)$ for $\z\sim p(\z|\x)$\footnote{Here, by a random predictor we mean that the predictor outputs a random vector from $\Delta^{k-1}$, not $\out$.}.
To incorporate the randomness of $\f$,
one only needs to change the expectation in \eqref{eq:mlls-main} to be over both $\x$ and $\f(\x)$,
then \eqref{eq:mlls-main} becomes a rewrite of \eqref{eq:mlls-z}. 
%
%

Proposition~\ref{prop:ltt-calib} indicates that constructing 
the confusion matrix is a calibration procedure.
Thus, the predictor constructed with constructed using confusion matrix is calibrated
and suitable for application with MLLS.
%
%
\begin{restatable}[\citet{vaicenavicius2019evaluating}]{reprop}{recalib}\label{prop:ltt-calib}
For any function $g$, $\f(\x) = p_s(y|g(\x))$ is a calibrated predictor.
\end{restatable}


We can now summarize the relationship between BBSE and MLLS:
A label shift estimator involves two design choices: 
(i) designing the latent space $p(\z|\x)$ 
(which is equivalent to designing a calibrated predictor); 
and (ii) performing distribution matching
in the new space $\ltt$. 
In BBSE, we design a calibrated predictor 
via the confusion matrix
and then perform distribution matching 
by directly solving linear equations. 
In general, MLLS does not specify 
how to obtain a calibrated predictor, 
but specifies KL minimization as 
the distribution matching procedure.
One can apply the confusion matrix approach 
to obtain a calibrated predictor 
and then plug it into MLLS, 
which is the BBSE analog under MLLS, 
and is a special case of MLLS.

\vspace{-8pt}
\section{Theoretical Analysis of MLLS}
\vspace{-5pt}
 \label{sec:finite}

We now analyze the performance of MLLS estimator. 
Even when $\w^*$ is the unique optimizer of \eqref{eq:mlls-main} for some calibrated predictor $\f$, assuming convex optimization can be done perfectly, there are still two sources of error preventing us from
exactly computing $\w^*$ in practice. 
First, we are optimizing a sample-based approximation \eqref{eq:mlls-sample} to the objective in expectation \eqref{eq:mlls-main}. 
We call this source of error \emph{finite-sample error}.  
Second, the predictor $\f$ we use may not be perfectly calibrated on the source distribution
%
as 
we only have access to samples from source data distribution $p_s(\x, y)$. 
We call this source of error \emph{miscalibration error}.
We will first analyze how these two sources of errors affect the estimate of $\w^*$ separately 
and then give a general error bound that incorporates both.
All proofs are relegated to Appendix~\ref{sec:AppendixB}.

Before presenting our analysis, 
we introduce 
some notation and regularity assumptions. 
For any predictor $\f:\inpt\mapsto \Delta^{k-1}$, 
we define $\w_{\f}$ and $\widehat{\w}_{\f}$ as in~\eqref{eq:mlls-main} and~\eqref{eq:mlls-sample}.
If $\f$ satisfies the conditions in Theorem~\ref{theorem:convergence} 
(calibration and linear independence) then we have that $\w_{\f} = \w^*$. 
Our goal is to bound $\norm{\widehat{\w}_{\f} - \w^*}$ 
for a given (possibly miscalibrated) predictor $\f$.
We now introduce a regularity condition:
\begin{condition}[Regularity condition for a predictor $\f$]
For any $\x$ within the support of $p_t(\x)$, i.e. $p_t(\x) > 0$, we have both $\f(\x)^T\w_{\f} \ge \tau$, $\f(\x)^T\w^* \ge \tau$ for some universal constant $\tau>0$.
\label{cond:tau}
\end{condition}
Condition~\ref{cond:tau} is mild if $\f$ is calibrated 
since in this case $\w_{\f} = \w^*$ 
is the maximizer of $\EE{t}{\log \f(\x)^T\w}$, and the condition is
satisfied if the expectation is finite. 
Since $\f(\x)^T\w^*$ and $\f(\x)^T\w_{\f}$ are upper-bounded (they are the inner products of two vectors which sum to 1),
they also must be lower-bounded away from $0$
with arbitrarily high probability without any assumptions. 
For miscalibrated $\f$, a similar justification 
holds for assumption that $\f(\x)^T\w_{\f}$ is lower bounded. 
Turning our attention to the assumption that $\f(\x)^T\w^*$ is lower bounded, we note that 
it is sufficient if $\f$ is close (pointwise) to some calibrated predictor.
This in turn is a reasonable assumption 
on the actual predictor we use for MLLS in practice 
as it is post-hoc calibrated on source data samples.

Define $\sigma_{\f, \w}$ to be the minimum eigenvalue 
of the Hessian $- \nabla_{\w}^2 \LL(\w, \f)$. To state our results compactly
we use standard stochastic order notation (see, for instance,~\citep{van1996weak}). 
We first bound the estimation error introduced
by only having finite samples from the target distribution in Lemma~\ref{lemma:bound-finite}. Next, we bound the estimation error introduced by having a miscalibrated $\f$ in Lemma~\ref{lemma:bound-calib}.

\begin{restatable}[]{relemma}{errone} \label{lemma:bound-finite}
For any predictor $\f$ that satisfies Condition~\ref{cond:tau}, we have 
$\norm{\w_{\f} - \widehat{\w}_{\f}} \le \sigma_{\f, \w_{\f}}^{-1}  \comp_p \left(  m^{-1/2} \right).$
\end{restatable}
\begin{restatable}[]{relemma}{errtwo}\label{lemma:bound-calib}
For any predictor $\f$ and any calibrated predictor $\f_c$ that satisfies Condition~\ref{cond:tau}, we have 
$\norm{\w_{\f} - \w^*} \le  \sigma_{\f, \w^*}^{-1} \cdot C\cdot \EE{t}{\norm{\f - \f_c}} \,,$
for some constant $C$.

If we set $\f_c(\x) = p_s(y | \f(\x))$, 
which is a calibrated predictor (Proposition~\ref{prop:ltt-calib}), 
we can bound the error in terms of the calibration error of $\f$ on the source data \footnote{We present two upper bounds because the second is more interpretable while the first is tighter.}:
$\norm{\w_{\f} - \w^*} \le \sigma_{\f, \w^*}^{-1} \cdot C\cdot\CE{\f} \,.$
\end{restatable}
%
Note that since $p_s(y)>0$ for all $y$, we can upper-bound the error 
in Lemma~\ref{lemma:bound-calib} with calibration error 
on the source data.
We combine the two sources of error to bound the estimation error $\norm{\widehat{\w}_{\f} - \w^*}$:
\begin{restatable}[]{rethm}{main-error-bound}\label{theorem:main-error-bound}
For any predictor $\f$ that satisfies Condition~\ref{cond:tau}, we have
\begin{align*}
\norm{\widehat{\w}_{\f} - \w^*} \le \sigma_{\f, \w_{\f}}^{-1} \comp_p \left(  m^{-1/2} \right) + C\cdot \sigma_{\f, \w^*}^{-1} \CE{\f} \,. \addeq\label{eq:bound-main} 
\end{align*}
\end{restatable}
The estimation error of MLLS can be decomposed into 
(i) finite-sample error, which decays at a rate of $m^{-1/2}$;
and (ii) the calibration error of the predictor that we use.
The proof is a direct combination of Lemma~\ref{lemma:bound-finite} 
and Lemma~\ref{lemma:bound-calib} applied to the same $\f$ with the following error decomposition:
\begin{align*}
\norm{\widehat{\w}_{\f} - \w^*} \le  \underbrace{ \norm{\w_{\f} - \widehat{\w}_{\f}}}_{\textrm{finite-sample}} +  \underbrace{\norm{\w_{\f} - \w^*}}_{\textrm{miscalibration}} \,.
\end{align*}
Theorem~\ref{theorem:main-error-bound} shows that 
the estimation error depends inversely 
on the minimum eigenvalue of the Hessian 
at two different points $\w_{\f}$ and $\w^*$. 
One can unify these two eigenvalues 
as a single quantity $\sigma_{\f}$, the minimum eigenvalue $\EE{t}{\f(\x)\f(\x)^T}$. We formalize this observation 
in Appendix~\ref{sec:AppendixB}. 




If we use the \emph{post-hoc calibration} procedure
(as discussed in Section~\ref{sec:pre} and~\ref{sec:MLLs_alg}) to calibrate a blackbox predictor $\smash{\widehat{\f}}$, 
we can obtain a bound on the calibration error of $\f$. In more detail, suppose that the class $\G$ 
used for calibration satisfies standard regularity conditions (injectivity, Lipschitz-continuity, twice differentiability, non-singular Hessian).
We have the following lemma:  
\begin{restatable}[]{relemma}{ce}\label{lemma:ce}
Let $\f = \g \circ \widehat{\f}$ be the predictor after post-hoc calibration 
with squared loss $l$ and $\g$ belongs to a function class $\G$ 
that satisfies the standard regularity conditions, we have 
\begin{align}\label{eq:platt}
\CE{\f} \le  \min_{\g \in \G} \CE{\g \circ \widehat{\f} } + \comp_p\left( n^{-1/2}\right) \,. 
\end{align}
\end{restatable}
%

This result is similar to Theorem 4.1~\cite{kumar2019verified}. 
For a model class $\G$ that is rich enough to contain a function $\g \in \G$ 
that achieves zero calibration error, i.e.,
$\min_{\g \in \G} \CE{\g \circ \widehat{\f} } = 0$, 
then we obtain an estimation error bound for MLLS 
of $\sigma_{\f}^{-1} \cdot\comp_p \left(  m^{-1/2} +  n^{-1/2}\right)$. 
This bound is similar to rate of RLLS and BBSE,
where instead of $\sigma_{\f}$ they have minimum eigenvalue of the confusion matrix.

%
%

%
%
The estimation error bound explains the efficiency of MLLS.
Informally, the error of MLLS depends inversely
on the minimum eigenvalue of the Hessian of the likelihood $\sigma_{\f}$.
When we apply coarse calibration via the confusion matrix (in MLLS-CM),
we only decrease the value of $\sigma_{\f}$.
Coarse calibration throws away information~\citep{kuleshov2015calibrated} and thus results 
in greater estimation error for MLLS. 
%
In Section~\ref{sec:exp}, we empricially show that MLLS-CM's performance
is similar to that of BBSE. 
Moreover, on a synthetic Gaussian mixture model, 
we show that the minimum eigenvalue of the Hessian obtained 
using confusion matrix calibration is smaller than the minimum eigenvalue obtained with more granular calibration. 
Our analysis and observations together suggest MLLS's superior performance than BBSE (or RLLS)
is due to the granular calibration but not 
due to the difference in the optimization objective. 
%

Finally, we want to highlight one minor point regarding applicability of our result.
If $\f$ is calibrated, Theorem~\ref{theorem:main-error-bound},
together with Proposition~\ref{prop:eigen} (in Appendix~\ref{sec:AppendixB}),
implies that MLLS is consistent if $\EE{t}{\f(\x)\f(\x)^T}$ is invertible. 
Compared to the consistency condition in Theorem~\ref{theorem:mlls} 
that $\EE{s}{\f(\x)\f(\x)^T}$ is invertible 
(together with Proposition~\ref{prop:equiv-cond}), 
these two conditions are the same if the likelihood ratio $p_t(\f(\x)) / p_s(\f(\x))$ is lower-bounded. 
This is true if all entries in $\w^*$ are non-zero. 
Even if $\w^*$ contains non-zero entries, 
the two conditions are still the same if there exists some $\w^*_y > 0$ 
such that $p(\f(\x)|y)$ covers the full support of $p_s(\f(\x))$. 
In general however, the invertibility of $\EE{t}{\f(\x)\f(\x)^T}$ 
is a stronger requirement than the invertibility of $\EE{s}{\f(\x)\f(\x)^T}$.
We leave further investigation of this gap for future work.





\vspace{-8pt}
\section{Experiments}
\vspace{-5pt}
	\label{sec:exp}

\begin{figure*}[t!] 
\centering
    \subfigure[GMM]{ \includegraphics[width=0.3\linewidth]{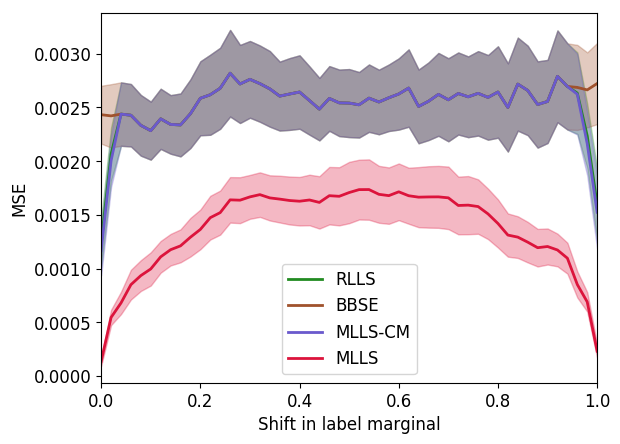}}\hfil
    \subfigure[MNIST]{\includegraphics[width=0.3\linewidth]{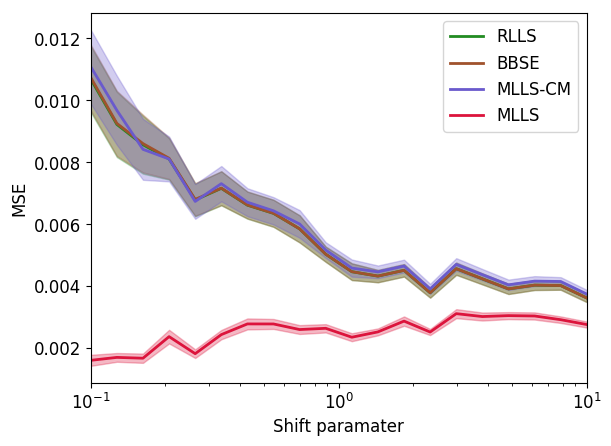}}\hfil
    \subfigure[CIFAR-10]{\includegraphics[width=0.3\linewidth]{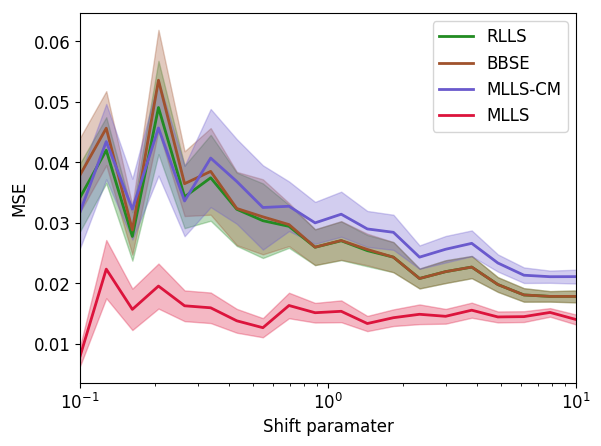}}\par\medskip
    \vspace{-10pt}
    \subfigure[GMM]{\includegraphics[width=0.3\linewidth]{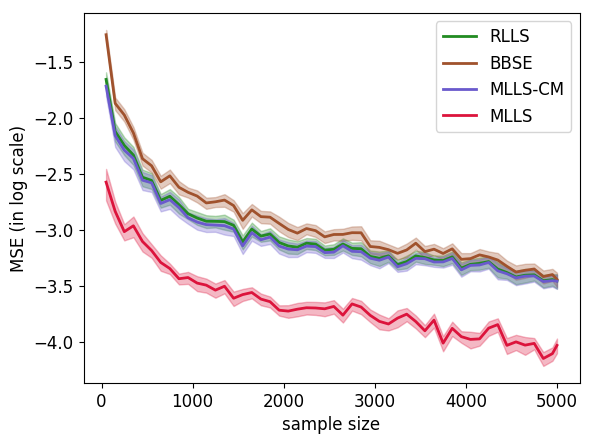}}\hfil
    \subfigure[MNIST]{\includegraphics[width=0.3\linewidth]{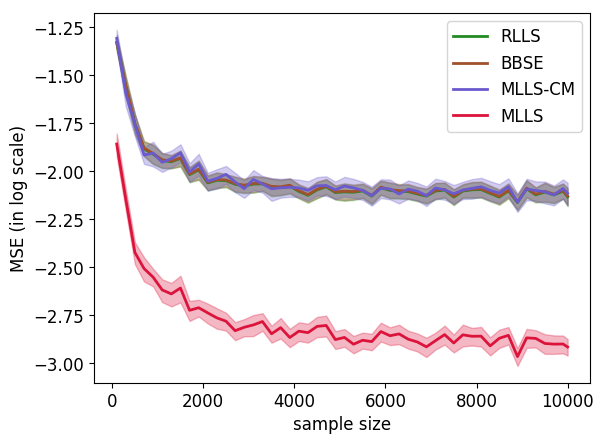}}\hfil
    \subfigure[CIFAR-10]{\includegraphics[width=0.3\linewidth]{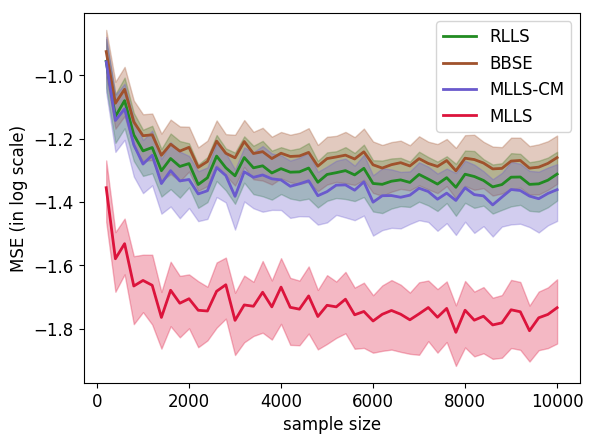}}
    \vspace{-2pt}
\caption{ (\textbf{top}) MSE vs the degree of shift; For GMM, we control the shift in the label marginal for class 1 with a fixed target sample size of 1000. For multiclass problems----MNIST and CIFAR-10, we control the Dirichlet shift parameter with a fixed sample size of 5000. (\textbf{bottom}) MSE (in log scale) vs target sample size; For GMM, we fix the label marginal for class 1 at 0.01 whereas for multiclass problems, MNIST and CIFAR-10, we fix the Dirichlet parameter to 0.1. In all plots, MLLS dominates other methods. All confusion matrix approaches perform similarly, indicating that the advantage of MLLS  comes from the choice of calibration but not the way of performing distribution matching. 
}\label{figure:MSE_err} 
\vspace{-13pt}
\end{figure*}

We experimentally illustrate the performance of MLLS on synthetic data, 
MNIST \citep{lecun1998mnist}, and CIFAR10 \citep{krizhevsky2009learning}. 
Following \citet{lipton2018detecting}, we experiment with \emph{Dirichlet shift} simulations.
On each run, we sample a target label distribution $\pt(y)$
from a Dirichlet with concentration parameter $\alpha$.
We then generate each target example by first sampling a label $y\sim \pt(y)$
and then sampling (with replacement) an example conditioned on that label .
%
%
Note that smaller values of alpha correspond to more severe shift.
In our experiments, the source label distribution is uniform.

First, we consider a mixture of two Gaussians (as in Example in Section~\ref{subsec:MLLS}) with $\mu=1$. 
With CIFAR10 and MNIST, we split the full training set into two subsets: 
train and valid, and use the provided test set as is.  
Then according to the label distribution,
we randomly sample with replacement train, valid, and test set 
from each of their respective pool to form the source and target set. 
To learn the black box predictor on real datasets, 
we use the same architecture as~\citet{lipton2018detecting} 
for MNIST, and for CIFAR10 we use ResNet-18~\citep{he2016deep} 
as in \citet{azizzadenesheli2019regularized}\footnote{We 
used open source implementation of ResNet-18 https://github.com/kuangliu/pytorch-cifar.}. 
For simulated data, we use the true $\ps(y|x)$ as our predictor function. 
For each experiment, we sample $100$ datasets 
for each shift parameter and evaluate the empirical MSE 
and variance of the estimated weights. 

We consider three sets of experiments: 
(1) MSE vs degree of target shift;
(2) MSE vs target sample sizes; 
and (3) MSE vs calibrated predictors on the source distribution. We refer to MLLS-CM as MLLS with hard confusion matrix calibration as in \eqref{eq:cm-log}. 
In our experiments, we compare MLLS estimator with BBSE, RLLS, and MLLS-CM. 
For RLLS and BBSE, we use the publicly available code \footnote{BBSE: https://github.com/zackchase/label\_shift, RLLS:  https://github.com/Angela0428/labelshift}. 
To post-hoc calibration, we use BCTS ~\citep{alexandari2019adapting} 
on the held-out validation set. 
Using the same validation set, we calculate 
the confusion matrix for BBSE, RLLS, and MLLS-CM. 

We examine the performance of various estimators across all three datasets 
for various target dataset sizes and shift magnitudes (Figure~\ref{figure:MSE_err}).
Across all shifts, MLLS (with BCTS-calibrated classifiers)
\emph{uniformly dominates} BBSE, RLLS, and MLLS-CM 
in terms of MSE (Figure~\ref{figure:MSE_err}).
Observe for severe shifts, MLLS is comparatively dominant.
As the available target data increased, 
all methods improve rapidly,
with MLLS outperforming all other methods by a significant margin. 
Moreover, MLLS's advantages grow more pronounced under extreme shifts. 
Notice MLLS-CM is roughly equivalent to BBSE across all settings of dataset, target size, and shift magnitude. 
This concludes MLLS's superior performance is not because of differences in loss function used for distribution matching but due to differences in the granularity of the predictions, caused by crude confusion matrix aggregation.  

\begin{wrapfigure}{r}{0.4\textwidth}
\vspace{-10pt}
  \centering 
  \includegraphics[width=.33\columnwidth]{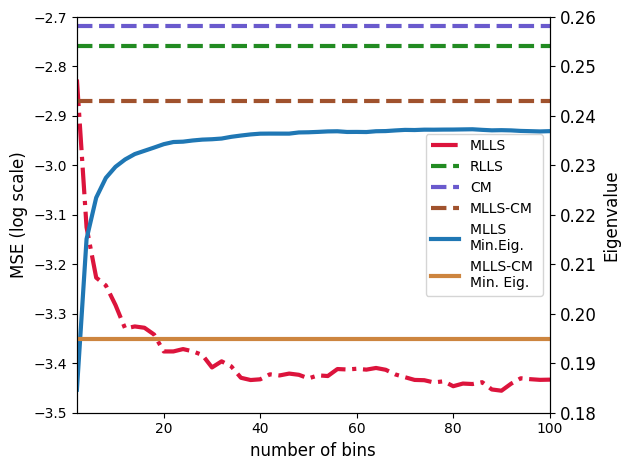} 
  \caption{ 
MSE (left-axis) with variation of minimum eigenvalue of the Hessian (right-axis) vs number of bins used for aggregation.  With increase in number of bins, MSE decrease and the minimum eigenvalue increases.}\label{fig:Gbin}
\vspace{-10pt}
\end{wrapfigure}

Note that given a predictor $\f_1$,
we can partition our input space and produce another predictor $\f_2$ that, 
for any data-point gives the expected output of $\f_1$ 
on points belonging to that partition.
If $\f_1$ is calibrated, then $\f_2$ will also 
be calibrated \citep{vaicenavicius2019evaluating}.
%
%
On synthetic data, we vary the granularity of calibration (for MLLS)
%
by aggregating $\ps(y|x)$ over a variable number of equal-sized bins. 
With more bins, less information is lost 
due to calibration.
Consequently, the minimum eigenvalue of the Hessian increases 
and the MSE decreases, supporting our theoretical bounds (Figure~\ref{fig:Gbin}).
We also verify that the confusion matrix calibration
performs poorly (Figure~\ref{fig:Gbin}).
For MLLS-CM, the minimum eigenvalue of the Hessian is 0.195, 
significantly smaller than for the binned predictor for $\#$bin $\ge4$. 
Thus, the poor performance of MLLS-CM is predicted by
its looser upper bound per our analysis.
%
Note that these experiments presume access to the true predictor $\ps(y|x)$ and thus the MSE strictly improves with the number of bins. 
In practice, with a fixed source dataset size,
increasing the number of bins could lead to overfitting, worsening our calibration.

\vspace{-12pt}
\section{Conclusion}
\vspace{-8pt}
This paper provides a unified framework relating techniques
that use off-the-shelf predictors for label shift estimation.
We argue that these methods all employ calibration,
either explicitly or implicitly, 
differing only in the choice of calibration method
and their optimization objective.
Moreover, with our analysis we show that the choice of calibration method
(and not the optimization objective for distribution matching) 
accounts for the advantage of MLLS with BCTS calibration 
over BBSE. 
In future work, we hope to operationalize these insights
to provide guidance for a calibration scheme 
to improve label shift estimation.

\newpage
\section*{Broader Impact}
This paper investigates the (statistical) consistency and efficiency of two existing methods for estimating target domain label distributions. While this could potentially guide practitioners to improve detection, estimation, and classification in applications where the label shift assumption holds, we do not believe that it will fundamentally impact how machine learning is used in a way that could conceivably be socially salient. While we take the potential impact of machine learning on society seriously, we believe that this work, which addresses a foundational theoretical problem, does not present a significant societal concern.

\section*{Acknowledgments}
We thank Zico Kolter and David Childers for their helpful feedback. 
This material is based on research sponsored by Air Force Research Laboratory (AFRL) under agreement number FA8750-19-1-1000. The U.S. Government is authorized to reproduce and distribute reprints for Government purposes notwithstanding any copyright notation therein. 
The views and conclusions contained herein are those of the authors and should not be interpreted as necessarily representing the official policies or endorsements, either expressed or implied, of Air Force Laboratory, DARPA or the U.S. Government. SB acknowledges funding from the NSF grants DMS-1713003 and CIF-1763734.
ZL acknowledges Amazon AI, Salesforce Research, Facebook, UPMC, Abridge, and the Center for Machine Learning and Health
for their generous support of ACMI Lab's research on machine learning under distribution shift.



\bibliographystyle{abbrvnat}
\bibliography{lifetime}

\begin{thebibliography}{28}
\providecommand{\natexlab}[1]{#1}
\providecommand{\url}[1]{\texttt{#1}}
\expandafter\ifx\csname urlstyle\endcsname\relax
  \providecommand{\doi}[1]{doi: #1}\else
  \providecommand{\doi}{doi: \begingroup \urlstyle{rm}\Url}\fi

\bibitem[Alexandari et~al.(2019)Alexandari, Kundaje, and
  Shrikumar]{alexandari2019adapting}
A.~Alexandari, A.~Kundaje, and A.~Shrikumar.
\newblock Adapting to label shift with bias-corrected calibration.
\newblock In \emph{arXiv preprint arXiv:1901.06852}, 2019.

\bibitem[Azizzadenesheli et~al.(2019)Azizzadenesheli, Liu, Yang, and
  Anandkumar]{azizzadenesheli2019regularized}
K.~Azizzadenesheli, A.~Liu, F.~Yang, and A.~Anandkumar.
\newblock Regularized learning for domain adaptation under label shifts.
\newblock In \emph{International Conference on Learning Representations
  (ICLR)}, 2019.

\bibitem[Ben-David et~al.(2010)Ben-David, Lu, Luu, and
  P{\'a}l]{ben2010impossibility}
S.~Ben-David, T.~Lu, T.~Luu, and D.~P{\'a}l.
\newblock {Impossibility Theorems for Domain Adaptation}.
\newblock In \emph{International Conference on Artificial Intelligence and
  Statistics (AISTATS)}, 2010.

\bibitem[Blanchard et~al.(2010)Blanchard, Lee, and Scott]{blanchard2010semi}
G.~Blanchard, G.~Lee, and C.~Scott.
\newblock Semi-supervised novelty detection.
\newblock \emph{Journal of Machine Learning Research}, 11\penalty0
  (Nov):\penalty0 2973--3009, 2010.

\bibitem[Byrd and Lipton(2019)]{byrd2019effect}
J.~Byrd and Z.~C. Lipton.
\newblock What is the effect of importance weighting in deep learning?
\newblock In \emph{International Conference on Machine Learning (ICML)}, 2019.

\bibitem[Cortes and Mohri(2014)]{cortes2014domain}
C.~Cortes and M.~Mohri.
\newblock Domain adaptation and sample bias correction theory and algorithm for
  regression.
\newblock \emph{Theoretical Computer Science}, 519, 2014.

\bibitem[Cortes et~al.(2010)Cortes, Mansour, and Mohri]{cortes2010learning}
C.~Cortes, Y.~Mansour, and M.~Mohri.
\newblock {Learning Bounds for Importance Weighting}.
\newblock In \emph{Advances in Neural Information Processing Systems (NIPS)},
  2010.

\bibitem[Du~Plessis and Sugiyama(2014)]{du2014semi}
M.~C. Du~Plessis and M.~Sugiyama.
\newblock Semi-supervised learning of class balance under class-prior change by
  distribution matching.
\newblock \emph{Neural Networks}, 50:\penalty0 110--119, 2014.

\bibitem[Gretton et~al.(2009)Gretton, Smola, Huang, Schmittfull, Borgwardt, and
  Sch{\"o}lkopf]{gretton2009covariate}
A.~Gretton, A.~J. Smola, J.~Huang, M.~Schmittfull, K.~M. Borgwardt, and
  B.~Sch{\"o}lkopf.
\newblock {Covariate Shift by Kernel Mean Matching}.
\newblock \emph{Journal of Machine Learning Research (JMLR)}, 2009.

\bibitem[Guo et~al.(2017)Guo, Pleiss, Sun, and Weinberger]{guo2017calibration}
C.~Guo, G.~Pleiss, Y.~Sun, and K.~Q. Weinberger.
\newblock On calibration of modern neural networks.
\newblock In \emph{International Conference on Machine Learning (ICML)}, 2017.

\bibitem[He et~al.(2016)He, Zhang, Ren, and Sun]{he2016deep}
K.~He, X.~Zhang, S.~Ren, and J.~Sun.
\newblock {Deep Residual Learning for Image Recognition}.
\newblock In \emph{Computer Vision and Pattern Recognition (CVPR)}, 2016.

\bibitem[Krizhevsky and Hinton(2009)]{krizhevsky2009learning}
A.~Krizhevsky and G.~Hinton.
\newblock {Learning Multiple Layers of Features from Tiny Images}.
\newblock Technical report, Citeseer, 2009.

\bibitem[Kuleshov and Liang(2015)]{kuleshov2015calibrated}
V.~Kuleshov and P.~S. Liang.
\newblock Calibrated structured prediction.
\newblock In \emph{Advances in Neural Information Processing Systems
  (NeurIPS)}, 2015.

\bibitem[Kumar et~al.(2019)Kumar, Liang, and Ma]{kumar2019verified}
A.~Kumar, P.~S. Liang, and T.~Ma.
\newblock Verified uncertainty calibration.
\newblock In \emph{Advances in Neural Information Processing Systems
  (NeurIPS)}, 2019.

\bibitem[LeCun et~al.(1998)LeCun, Bottou, Bengio, and Haffner]{lecun1998mnist}
Y.~LeCun, L.~Bottou, Y.~Bengio, and P.~Haffner.
\newblock {Gradient-Based Learning Applied to Document Recognition}.
\newblock \emph{Proceedings of the IEEE}, 86, 1998.

\bibitem[Lipton et~al.(2018)Lipton, Wang, and Smola]{lipton2018detecting}
Z.~C. Lipton, Y.-X. Wang, and A.~Smola.
\newblock {Detecting and Correcting for Label Shift with Black Box Predictors}.
\newblock In \emph{International Conference on Machine Learning (ICML)}, 2018.

\bibitem[Ramaswamy et~al.(2016)Ramaswamy, Scott, and
  Tewari]{ramaswamy2016mixture}
H.~Ramaswamy, C.~Scott, and A.~Tewari.
\newblock Mixture proportion estimation via kernel embeddings of distributions.
\newblock In \emph{International Conference on Machine Learning}, pages
  2052--2060, 2016.

\bibitem[Saerens et~al.(2002)Saerens, Latinne, and
  Decaestecker]{saerens2002adjusting}
M.~Saerens, P.~Latinne, and C.~Decaestecker.
\newblock {Adjusting the Outputs of a Classifier to New a Priori Probabilities:
  A Simple Procedure}.
\newblock \emph{Neural Computation}, 2002.

\bibitem[Sch{\"o}lkopf et~al.(2012)Sch{\"o}lkopf, Janzing, Peters, Sgouritsa,
  Zhang, and Mooij]{scholkopf2012causal}
B.~Sch{\"o}lkopf, D.~Janzing, J.~Peters, E.~Sgouritsa, K.~Zhang, and J.~Mooij.
\newblock {On Causal and Anticausal Learning}.
\newblock In \emph{International Conference on Machine Learning (ICML)}, 2012.

\bibitem[Shimodaira(2000)]{shimodaira2000improving}
H.~Shimodaira.
\newblock {Improving Predictive Inference Under Covariate Shift by Weighting
  the Log-Likelihood Function}.
\newblock \emph{Journal of Statistical Planning and Inference}, 2000.

\bibitem[Stein(1981)]{stein1981estimation}
C.~M. Stein.
\newblock Estimation of the mean of a multivariate normal distribution.
\newblock \emph{The annals of Statistics}, pages 1135--1151, 1981.

\bibitem[Storkey(2009)]{storkey2009training}
A.~Storkey.
\newblock {When Training and Test Sets Are Different: Characterizing Learning
  Transfer}.
\newblock \emph{Dataset Shift in Machine Learning}, 2009.

\bibitem[Tropp et~al.(2015)]{tropp2015introduction}
J.~A. Tropp et~al.
\newblock An introduction to matrix concentration inequalities.
\newblock \emph{Foundations and Trends{\textregistered} in Machine Learning},
  2015.

\bibitem[Vaicenavicius et~al.(2019)Vaicenavicius, Widmann, Andersson, Lindsten,
  Roll, and Sch{\"o}n]{vaicenavicius2019evaluating}
J.~Vaicenavicius, D.~Widmann, C.~Andersson, F.~Lindsten, J.~Roll, and T.~B.
  Sch{\"o}n.
\newblock Evaluating model calibration in classification.
\newblock In \emph{International Conference on Machine Learning (ICML)}, 2019.

\bibitem[van~de Geer(2000)]{geer2000empirical}
S.~van~de Geer.
\newblock \emph{Empirical Processes in M-estimation}, volume~6.
\newblock Cambridge university press, 2000.

\bibitem[van~der Vaart and Wellner(1996)]{van1996weak}
A.~W. van~der Vaart and J.~A. Wellner.
\newblock Weak convergence.
\newblock In \emph{Weak convergence and empirical processes}. Springer, 1996.

\bibitem[Zadrozny(2004)]{zadrozny2004learning}
B.~Zadrozny.
\newblock {Learning and Evaluating Classifiers Under Sample Selection Bias}.
\newblock In \emph{International Conference on Machine Learning (ICML)}, 2004.

\bibitem[Zhang et~al.(2013)Zhang, Sch{\"o}lkopf, Muandet, and
  Wang]{zhang2013domain}
K.~Zhang, B.~Sch{\"o}lkopf, K.~Muandet, and Z.~Wang.
\newblock {Domain Adaptation Under Target and Conditional Shift}.
\newblock In \emph{International Conference on Machine Learning (ICML)}, 2013.

\end{thebibliography}

\clearpage

\appendix



%


\section{MLLS Algorithm} \label{sec:MLLs_alg}

\begin{algorithm}
  \caption{Maximum Likelihood Label Shift estimation}
  \label{alg:MLLS}
\begin{algorithmic}[1]
  \INPUT: Labeled validation samples from source and unlabeled test samples from target. Trained blackbox model $\hat{\f}$, model class $\G$ and loss function $l$ for calibration (for instance, MSE or negative log-likelihood).
    \STATE On validation data minimize the loss $l$ over class $\G$ to obtain $\f = \g \circ \hat{\f}$. 
  \STATE Solve the optimization problem \eqref{eq:mlls-sample} using $\f$ to get $\widehat \w$. 
  \OUTPUT: MLLS estimate $\widehat \w$ 
\end{algorithmic}
\end{algorithm}


\textbf{Step 1. description.}    
Let the model class used for \emph{post-hoc calibration} be represented by $\G$. 
Given a validation dataset $\{(x_{v1},y_{v1}),\ldots,(x_{vn},y_{vn})\}$ sampled from the 
source distribution $P_s$ we compute,
$\{(\widehat\f(\x_{v1}),y_{v1}), (\widehat\f(\x_{v2}), y_{v2}), \ldots , (\widehat\f(\x_{vn}), y_{vn})\}$, applying our
classifier $\widehat{f}$ to the data.
Using this we estimate
a function, 
\begin{align}
\widehat \g = \argmin_{\g \in \G} \sum_{i=1}^n \ell(\g \circ \widehat \f(\x_{vi}), y_{vi}) \,, \label{eq:posthoc}
\end{align}
where the loss function $\ell$ 
can be the negative log-likelihood or squared error. Experimentally we observe same performance with both the loss functions.  
Subsequently, we can apply the calibrated predictor $\widehat{g} \circ \widehat{f}$. 

%
Our experiments follow \citet{alexandari2019adapting},
who leverage BCTS \footnote{Motivated by the strong empirical results in \citet{alexandari2019adapting}, 
we use BCTS in our experiments as a surrogate for canonical calibration.}
to calibrate their models. 
BCTS extends temperature scaling~\citep{guo2017calibration} by incorporating 
per-class bias terms. 
Formally, a function $g:\Delta^{k-1} \mapsto \Delta^{k-1}$ in the BCTS class $\G$, is given by
\begin{align*}
g_j(x) = \frac{ \exp \left[ \log (x_j)/T + b_j\right] }{\sum_i \exp \left[ \log (x_i)/T + b_i \right ]} \quad \forall j \in \out \, 
\end{align*}
where $\{T, b_1,\ldots, b_{\abs{\out}}\}$
are the $\abs{\out}+1$ parameters to be learned.

\section{Prior Work on Label Shift Estimation} \label{sec:rel_work}

Dataset shifts are predominantly studied under two scenarios: covariate shift and label shift~\cite{storkey2009training}. \citet{scholkopf2012causal} articulates connections between label shift and covariate shift with anti-causal and causal models respectively. Covariate shift is well explored in past \cite{zhang2013domain,zadrozny2004learning,cortes2010learning,cortes2014domain,gretton2009covariate}. 

Approaches for estimating label shift (or prior shift) can be categorized into three classes: \begin{enumerate}
    \item Methods that leverage Mixture Proportion Estimation (MPE)~\citep{blanchard2010semi, ramaswamy2016mixture} techniques to estimate the target label distribution. MPE estimate in general (e.g. \citet{blanchard2010semi}) needs explicit calculations of $p_s(x|y) (= p_t(x|y))$ which is infeasible for high dimensional data. More recent methods for MPE estimation, i.e. \citet{ramaswamy2016mixture}, uses Kernel embeddings, which like many kernel methods, require the inversion of an $n \times n$ Gram matrix.  The $\mathcal{O}(n^3)$ complexity makes them infeasible for large datasets, practically used in deep learning these days;
    \item Methods that directly operate in RKHS for distribution matching~\citep{zhang2013domain,du2014semi}. \citet{zhang2013domain} extend the kernel mean matching approach due to \citet{gretton2009covariate} to the label shift problem. Instead of minimizing maximum mean discrepancy, \citet{du2014semi} explored minimizing PE divergence between the kernel embeddings to estimate the target label distribution. Again, both the methods involve inversion of an $n\times n$ kernel matrix, rendering them infeasible for large datasets; and
    \item  Methods that work in low dimensional setting ~\citep{lipton2018detecting,azizzadenesheli2019regularized,saerens2002adjusting} by directly estimating $\pt(y)/\ps(y)$ to avoid the curse of dimensionality. These methods leverage an off-the-shelf predictor to estimate the label shift ratio.
\end{enumerate}

In this paper, we primarily focus on unifying methods that fall into the third category. 

\section{Marginal calibration is insufficient to achieve consistency} \label{ref:marg_calib}

In this section, we will illustrate insufficiency of \emph{marginal calibration} to achieve consistency. For completeness, we first define margin calibration: 
\begin{definition}[Marginal calibration]~\label{def:mcalib}
 A prediction model $\f: \mathcal{X} \mapsto \Delta^{k-1}$ is marginally calibrated  
 on the source domain 
 if for all $\x \in \inpt$ and $j \in \out$, 
 \begin{align*}
 \mathrm{P}_s ( y = j | \f_j(x) ) = \f_j(x) \,.
 \end{align*}
\end{definition}

Intuitively, this definition captures per-label calibration of the classifier which is strictly less restrictive than requiring canonical calibration. In the example, we construct a classifier on discrete $\inpt$ which is marginally calibrated, but not canonically calibrated. With the constructed example, we show that the population objective~\eqref{eq:mlls-main} yields inconsistent estimates. 

\textbf{Example}. Assume $\inpt = \{x_1, x_2, x_3, x_4, x_5, x_6\}$ and $\out = \{1,2,3\}$. Suppose the predictor $\f(x)$ and $P_s(y | \f(x))$ are given as, 
    
    \qquad \qquad \begin{tabular}{ c || c | r | r }
      $\f(x)$ & y=1 & y=2 & y=3 \\
      \hline
      \hline
      $x_1$ & 0.1 & 0.2 & 0.7 \\
      \hline
      $x_2$ & 0.1 & 0.7 & 0.2 \\
      \hline
      $x_3$ & 0.2 & 0.1 & 0.7 \\
      \hline
      $x_4$ & 0.2 & 0.7 & 0.1 \\
      \hline
      $x_5$ & 0.7 & 0.1 & 0.2 \\
      \hline
      $x_6$ & 0.7 & 0.2 & 0.1 \\
    \end{tabular} \qquad \qquad
    \begin{tabular}{ c || c | r | r }
      $P_s(y | \f(x))$ & y=1 & y=2 & y=3 \\
      \hline
      \hline
      $x_1$ & 0.2 & 0.1 & 0.7 \\
      \hline
      $x_2$ & 0.0 & 0.8 & 0.2 \\
      \hline
      $x_3$ & 0.1 & 0.2 & 0.7 \\
      \hline
      $x_4$ & 0.3 & 0.6 & 0.1 \\
      \hline
      $x_5$ & 0.8 & 0.0 & 0.2 \\
      \hline
      $x_6$ & 0.6 & 0.3 & 0.1 \\
    \end{tabular}

Clearly, the prediction $f(x)$ is marginally calibrated. We have one more degree to freedom to choose, which is the source marginal distribution on $\inpt$. For simplicity let's assume $\ps(x_i) = 1/6$ for all $i = \{1, \ldots, 6\}$. Thus, we have $\ps(y=j) = 1/3$ for all $j = \{1,2,3\}$. Note, with our assumption of the source marginal on x, we get $P_t(x_i | y=j) = P_s(x_i | y=j) = P_s(y=j | \f(x_i))/2$. This follows as $x \mapsto \f(x)$ is an one-to-one mapping.

Now, assume a shift i.e. prior on $\out$ for the target distribution of the form $[\alpha, \beta, 1- \alpha - \beta]$.
With the label shift assumption, we get 
$$\forall i \qquad \pt(x_i) = \frac{1}{2} \left(\alpha P_s(y=1|f(x_i)) + \beta P_s(y=2|f(x_i)) + (1- \beta -\alpha) P_s(y=3|f(x_i))\right) \,.$$ 

Assume the importance weight vector as $w$. Clearly, we have $w_1 + w_2 + w_3 = 3$. 
Re-writing the population MLLS objective~\eqref{eq:mlls-main}, we get the maximisation problem as
\begin{align*}
    \argmax_w \sum_{i=1}^6 \pt(x_i) \log (\f(x_i)^T w) \,. \numberthis \label{eq:max}
\end{align*}  

Differentiating~\eqref{eq:max} with respect to $w_1$ and $w_2$, we get two high order equations, solving which give us the MLLS estimate $w_f$. 
To show inconsistency, it is enough to consider one instantiation of $\alpha$ and $\beta$ such that $\abs{3\alpha - w_1} + \abs{3\beta - w_2} + \abs{ w_1 + w_2 - 3\alpha - 3\beta} \ne 0$. Assuming $\alpha = 0.8$ and $\beta=0.1$ and  solving~\eqref{eq:max} using  numerical methods, we get $w_f = [2.505893, 0.240644, 0.253463]$. As $w  = [2.4, 0.3, 0.3 ]$, we have $w_f \ne w$ concluding the proof.

\section{Proofs from Section~\ref{sec:unify}} 
\label{sec:AppendixA}

\identifiability*

\begin{proof}
First we prove sufficiency. 
If there exists $\w\ne \w^*$ such that \eqref{eq:distr-match} holds,
then we have $\sum_{y=1}^k p_s(\z, y)(\w_y-\w^*_y)=0$ for all $\z \in \ltt$.
As $\w-\w^*$ is not the zero vector, 
$\{p_s(\z, y), y=1,...,k\}$ are linearly dependent. 
Since $p_s(\z, y)=p_s(y)p(\z|y)$ and $p_s(y)>0$ for all $y$ (by assumption), 
we also have that $\{p(\z|y), y=1,...,k\}$ are linearly dependent.
By contradiction, we show that the linear independence is necessary.

To show necessity, assume $\w^*_y = \frac{1}{kp_s(y)}$ for $y=1,...,k$.
We know that $\w^*$ satisfies \eqref{eq:distr-match} by definition.
If linear independence does not hold, then there exists a vector $\vv\in \Real^k$ 
such that $\vv\ne 0$ and $\sum_{y=1}^k p_s(\z, y)\vv_y=0$ for all $\z\in \ltt$.
Since the $\w^*$ we construct is not on the boundary of $\W$,
we can scale $\vv$ such that $\w^*+ \alpha \vv \in \W$ where $\alpha \ge 0$ and  $\vv \ne 0$.
Therefore, setting $\w = \w^*+ \alpha \vv$ gives another solution for \eqref{eq:distr-match}, 
which concludes the proof.
\end{proof}

\calibobj*

\begin{proof}
The proof follows a sequence of straightforward manipulations. In more detail, 
\begin{align*}
\EE{t}{ \log \f(\x)^T  \w}
& = \int  \, p_t(\x) \log [\f(\x)^T \w] d\x \\
& = \int \int  p_t(\x) p(\z|\x) \log [\f(\x)^T \w] d\x d\z \\
& = \int \int p_t(\x) p(\z|\x) \I{\f(\x)=\z} \log [\f(\x)^T \w]  d\x d\z\\
& = \int \int  p_t(\x) p(\z|\x) \log [\z^T \w] d\x d\z \\
& = \int  p_t(\z) \log [\z^T \w]   d\z\\
& = \int p_t(\z) \log \Big[\sum_{y=1}^k p_s(y|\z) \w\Big] d\z \,,
\end{align*}
where the final step uses the fact that $f$ is calibrated.

\end{proof}

\consistency*

\begin{proof}
According to Lemma~\ref{lemma:calib} we know that maximizing \eqref{eq:mlls-main} is
the same as maximizing \eqref{eq:mlls-z} with $p(\z|\x)=\delta_{\f(\x)}$,
thus also the same as minimizing the KL divergence between $p_t(\z)$ and $p_{\w}(\z)$.
Since $p_t(\z) \equiv p_{\w^*}(\z)$ 
we know that $\w^*$ is a minimizer 
of the KL divergence such that the KL divergence is 0. 
We also have that $\mathrm{KL}(p_t(\z), p_{\w}(\z))=0$ 
if and only if $p_t(\z) \equiv p_{\w}(\z)$, 
so all maximizers of \eqref{eq:mlls-main} should satisfy \eqref{eq:distr-match}. 
According to Lemma~\ref{lemma:identi-1}, 
if the strict linear independence holds, 
then $\w^*$ is the unique solution of \eqref{eq:distr-match}.
Thus $\w^*$ is the unique maximizer of  \eqref{eq:mlls-main}.

\end{proof}

\equivalent*

\begin{proof}
We first show the equivalence of (1) and (2). 
If $\f$ is calibrated, we have $p_s(\f(\x))\f_y(\x) = p_s(y)p(\f(\x)|y)$ for any $\x, y$. 
Then for any vector $\vv \in \Real^k$ we have 
\begin{align*}
\sum_{y=1}^k \vv_y p(\f(\x)|y) = \sum_{y=1}^k \frac{\vv_y}{p_s(y)} p_s(y)p(\f(\x)|y) =  \sum_{y=1}^k \frac{\vv_y}{p_s(y)} p_s(\f(\x))\f_y(\x) = p_s(\f(\x)) \sum_{y=1}^k \frac{\vv_y}{p_s(y)} \f_y(\x) \,.
\addeq\label{eq:eqv-proof-1}
\end{align*}
On the other hand, we can have
\begin{align*}
\EE{s}{\f(\x)\f(\x)^T} = \int \f(\x)\f(\x)^T p_s( \f(\x)) d (\f(\x)) \,.
\addeq\label{eq:eqv-proof-2}
\end{align*}

If $\{p(\f(\x)|y)\,:\,y=1,\ldots,k\}$ are linearly dependent, 
then there exist $\vv\ne0$ such that \eqref{eq:eqv-proof-1} is zero for any $\x$. 
Consequently, there exists a non-zero vector $\uu$ with $\uu_y = \vv_y / p_s(y)$
such that $\uu^T\f(\x)=0$ for any $\x$ satisfying $p_s(\f(\x))>0$, 
which means $\uu^T\EE{s}{\f(\x)\f(\x)^T} \uu = 0$
and thus $\EE{s}{\f(\x)\f(\x)^T}$ is not invertible. 
On the other hand, if  $\EE{s}{\f(\x)\f(\x)^T}$ is non-invertible, 
then there exist some $\uu \ne 0$ such that $\uu^T\EE{s}{\f(\x)\f(\x)^T} \uu = 0$. Further as $\uu^T\EE{s}{\f(\x)\f(\x)^T} \uu = \int \uu^T\f(\x)\f(\x)^T\uu\ p_s(\x) d\x  =  \int \abs{\f(\x)^T\uu} p_s(\x) d\x $. 
As a result, the vector $\vv$ with $\vv_y = p_s(y) \uu_y$ satisfies 
that \eqref{eq:eqv-proof-1} is zero for any $\x$, which means 
$\{p(\f(\x)|y)\,:\,y=1,\ldots,k\}$ are not strictly linearly independent.


Let $\C$ be the soft confusion matrix of $\f$, then 
\begin{align*}
\C_{ij}= p_s(\widehat{y}=i, y=j)
& =  \int d(\f(\x)) \, \f_i(\x) p(\f(\x)|y=j) p_s(y=j) \\
& =  \int \, \f_i(\x) \f_j(\x) p_s( \f(\x)) d (\f(\x)) \,.
\end{align*}
Therefore, we have $\C = \EE{s}{\f(\x)\f(\x)^T}$, which means (2) and (3) are equivalent.

\end{proof}

We introduce some notation before proving consistency.
Let $\calP = \left\{ \inner{\f}{\w} | \w \in \W \right\}$ be the class of densities\footnote{Note that 
we use the term \emph{density} loosely here for convenience.
The actual density is $\inner{\f(\x)}{\w}\cdot \ps(\x)$ 
but we can ignore $\ps(\x)$ because it does not depend on our parameters.} 
for a given calibrated predictor $\f$.
Suppose $\widehat p_n, p_0 \in \calP$ are densities corresponding 
to MLE estimate and true weights, respectively. 
We use $h(p_1, p_2)$ to denote the Hellinger distance 
and $\text{TV}(p_1, p_2)$ to denote the total variation distance 
between two densities $p_1, p_2$.
$H_r(\delta, \calP , P)$ denotes $\delta$-entropy
for class $\calP$ with respect to metric $L_r(P)$.
Similarly, $H_{r,B}(\delta, \calP , P)$ denotes the corresponding bracketing entropy. 
Moreover, $P_n$ denotes the empirical random distribution 
that puts uniform mass on observed samples $\x_1, \x_2, \ldots \x_n$. 
Before proving consistency we need to re-state two results:  

\begin{restatable}[Lemma 2.1 \cite{geer2000empirical}]{relemma}{bentropy} \label{lemma:bentropy}
If P is a probability measure, for all $1\le r < \infty$, we have 
\[
H_{r,B} (\delta, \calG , P) \le H_{\infty} (\delta/2, \calG) \qquad \text{for all } \delta >0\,. 
\]
\end{restatable}

\begin{restatable}[Corollary 2.7.10 \cite{van1996weak}]{relemma}{lipschitzentropy} \label{lemma:lipschitzentropy}
Let $\mathcal{F}$ be the class of convex functions $f: C \mapsto [0,1]$ defined on a compact, convex set $C \subset \Real^d$ such that $\abs{f(x) - f(y)} \le L \norm{x-y}$ for every x,y. Then 
\[
H_{\infty} (\delta, \mathcal{F}) \le K \left( \frac{L}{\delta} \right)^{d/2} \,, 
\]
for a constant K that depends on the dimension $d$ and $C$. 
\end{restatable}

We can now present our proof of consistency,
which is based on Theorem 4.6 from~\citet{geer2000empirical}:

\begin{restatable}[Theorem 4.6 \cite{geer2000empirical}]{relemma}{vandegeer} \label{lemma:vandegeer}
Let $\calP$ be convex and define class $\calG = \left\{ \frac{2p}{p+p_0} | p \in \calP \right\}$. If 
\[ \frac{1}{n} H_1 (\delta, \calG , P_n ) \to_P 0 \,, \addeq \label{eq:vandegeer}\]
then $h(\widehat p_n, p_0) \to 0 $ almost surely. 
\end{restatable}

\convergence*
\begin{proof}

Assume 
the maximizer of~\eqref{eq:mlls-sample} is $\widehat \w_f$ and $p_0 = \inner{\f}{\w^*}$. 
Define class $\calG = \left\{ \frac{2p}{p+p_0} | p \in \calP \right\}$. To prove consistency, we first bound the bracketing entropy for class $\calG$ using Lemma~\ref{lemma:bentropy} and Lemma~\ref{lemma:lipschitzentropy}. 

Clearly $\calP$ is linear in parameters and hence, convex. 
Gradient of function $g\in \calG$ is given by $\frac{2p_0}{(p+p_0)^2}$ which in turn is bounded by $\frac{2}{p_0}$.  
Under assumptions of Condition~\ref{cond:tau},
the functions in $\calG$ are Lipschitz with constant $2/\tau$. We can bound the bracketing entropy $H_{2,B} (\delta, \calG , P )$ using Lemma~\ref{lemma:lipschitzentropy} and Lemma~\ref{lemma:bentropy} as  
$$H_{2,B} (\delta, \calG , P) \le H_{\infty} (\delta, \calG) \le K_1 \left(\frac{1}{\delta \tau} \right)^{k/2} \,,$$ 
for some constant $K_1$ that depends on $k$. 

On the other hand, for cases where $p_0$ can be arbitrarily close to zero, i.e., Condition~\ref{cond:tau} doesn't hold true,
we define $\tau (\delta)$ and $\calG_\tau$ as
\[ \tau(\delta) = \sup \left\{ \tau \ge 0 \ | \int_{p_0 \le \tau} p_0 d\x \le \delta^2 \right\} \,, \numberthis \label{eq:delta_ball} 
\]
\[
\calG_\tau = \left\{\frac{2p}{p+p_0} \I{p_0\ge \tau} \ |\  p \in \calP  \right\} \,.
\]
Using triangle inequality, for any $g_1, g_2 \in \calG$, we have 
\begin{align*}
\int \norm{g_1 - g_2}^2 d\x  &\le  \int \norm{g_1 - g_2}^2 \I{p_0\le \tau} d\x + \int \norm{g_1 - g_2}^2 \I{p_0\ge \tau} d\x  \\ 
&\le 2 \int \I{p_0\le \tau} d\x + \int \norm{g_1 - g_2}^2 \I{p_0\ge \tau} d\x \,. \numberthis \label{eq:traingle_ineq} 
\end{align*}
Assume $\tau(\delta)$ such that \eqref{eq:delta_ball} is satisfied. Using \eqref{eq:traingle_ineq}, we have 
$$ H_{2,B} (\delta, \calG , P ) \le H_{2,B} (\sqrt{3} \delta, \calG_{\tau(\delta)} , P )  \,.$$
Thus, for the cases where $p_0$ can be arbitrarily close to zero,
instead of bounding $ H_{2,B} (\delta, \calG , P ) $, we 
we bound $H_B(\delta, \calG_{\tau(\delta)} , P )$. 
For any $\delta >0$, there is a compact subset $K_{\delta} \in \inpt$,
such that $\ps(X\setminus K_{\delta}) < \delta $. 
Using arguments similar to above, function $g \in \calG_{\tau (\delta)}$ is Lipschitz 
with constant $2/\tau(\delta) > 0$.
Again using Lemma~\ref{lemma:lipschitzentropy} and Lemma~\ref{lemma:bentropy}, we conclude 
$$H_{2,B} (2\delta, \calG_{\tau(\delta)} , P) \le H_{\infty} (\delta, \calG_{\tau(\delta)}) \le K_2 \left( \frac{1}{\delta \tau(\delta)} \right)^k \,,$$ 
for some constant $K_2$ that depends on $k$. 
Finally, we use Lemma~\ref{lemma:vandegeer} to conclude 
$h(\widehat p_n, p_0) \to_{\text{a.s.}} 0$. 
Further, as $\text{TV}(\widehat p_n, p_0) \le h(\widehat p_n, p_0)$, 
we have $h(\widehat p_n, p_0) \to_{\text{a.s.}} 0$ implies 
$\text{TV}(\widehat p_n, p_0)\to_{\text{a.s.}} 0$.
Further

\begin{align*}
\norm{\wf - \w^*}^2 &\le \frac{1}{\lambda_{\min}} \int \abs{\f(\x)^T (\wf - \w^*)}^2 \ps (x) d\x \\
&\le \frac{\sup_\x \left\{\abs{\f(\x)^T (\wf - \w^*)}\right \}}{\lambda_{\min}} \underbrace{\int \abs{\f(\x)^T (\wf - \w^*)} \ps(\x) d\x}_{\text{TV}(\widehat p_n, p_0)} \,, \numberthis \label{eq:norm-bound}
\end{align*}
where $\lambda_{\min}$ is the minimum eigenvalue of covariance matrix $\left[ \int \f(\x) \f(\x)^T \ps (x) d\x \right]$. Note using Proposition~\ref{prop:equiv-cond}, we have $\lambda_{\min} > 0$.
Thus, we conclude $\norm{\widehat \w_f - \w^*} \to_{\text{a.s.}} 0$. 
\end{proof}


\textbf{Example 1.} 
Consider a mixture of two Gaussians with 
\allowdisplaybreaks{$\ps(x|y=0) \defeq \N(\mu, 1)$ 
and $\ps(x|y=1) \defeq \N(-\mu, 1)$}. 
We suppose that the source mixing coefficients are both $\frac{1}{2}$, while
the target mixing coefficients are $\alpha (\ne \frac{1}{2}), 1-\alpha $. 
Assume a class of probabilistic threshold classifiers:
$f(x) = [1-c, c]$ for $x\ge 0$, otherwise $f(x) = [c, 1-c]$ with $c \in [0,1]$.

Then the population error of MLLS is given by 
\begin{align*}
4\abs{\frac{(1-2\alpha)(\ps(x\ge0|y=0) - c)}{1-2c}},
\end{align*}
which is zero only if $c = \ps(x\ge0|y=0)$ for a non-degenerate classifier.

\begin{proof}
The intuition behind the construction is, for such an Example, we can get a closed form solution for the population MLLS and hence allows a careful analysis of the estimation error.
The classifier $f(x)$ predicts class $0$ with probability $c$
and class $1$ with probability $1-c$ for $x\ge0$, and vice-versa for $x<0$. 
Using such a classifier, the weight estimator is given by: 
\begin{align*}    
 \widehat w &= \argmin_w \E{ \log\inner{f(x)}{w}}  \\
&\stackrel{\text{(i)}}{=} \argmin_{w_0} \left[ \int_{-\infty}^{0} \log((1-c)w_0 + c(2-w_0)) \pt(x) dx + \int_{0}^{\infty} \log(cw_0 + (1-c)(2-w_0)) \pt(x) dx \right]\\
&\stackrel{\text{(ii)}}{=} \argmin_{w_0} \left[  \log((1-c)w_0 + c(2-w_0)) \pt(x\le 0)  + \log(cw_0 + (1-c)(2-w_0)) \pt(x\ge 0) \right] \,,
\end{align*}
where equality (i) follows from $w_1 = 2-w_0$ and the predictor function
and (ii) follows from the fact that within each integral, 
the term inside the log is independent of $x$.  
Differentiating w.r.t. to $w_0$, we have: 
$$ \frac{1-2c}{2c + w_0 - 2cw_0} \pt(x\le 0) + \frac{2c-1}{2cw_0 + 2 -2c -w_0}\pt(x\ge 0) = 0  $$
$$ \frac{1}{2c + w_0 - 2cw_0} \pt(x\le 0) + \frac{-1}{2cw_0 + 2 -2c -w_0}(1 - \pt(x\le 0)) = 0  $$
$$ (2cw_0 + 2 -2c -w_0) \pt(x\le 0) - (2c + w_0 - 2cw_0)(1 - \pt(x\le 0)) = 0 $$
$$ 2\pt(x\le 0) - 2c - w_0 +2cw_0 = 0 \,,$$
which gives $w_0 = \frac{2\pt(x\le 0) - 2c}{1-2c}$. 
Thus for the population MLLS estimate, 
the estimation error is given by
$$ \norm{ \widehat \w - \w^*} = 2 |w_0 - 2\alpha| = 4\abs{\frac{(1-2\alpha)(\ps(x\ge0|y=0) - c)}{1-2c}} \,. $$
\end{proof}

\section{Proofs from Section ~\ref{sec:finite}} \label{sec:AppendixB}

The gradient of the MLLS objective can be written as
\begin{align*}
\nabla_{\w} \LL(\w, \f) = \EE{t}{\frac{\f(\x)}{\f(\x)^T \w}} \,,
\addeq\label{eq:proof-gradient}
\end{align*}
and the Hessian is
\begin{align*}
\nabla^2_{\w} \LL(\w, \f) = - \EE{t}{\frac{\f(\x)\f(\x)^T}{\left(\f(\x)^T \w\right)^2}} \,.
\addeq\label{eq:proof-hessian}
\end{align*}
We use $\lambda_{\min}(X)$ to denote the minimum eigenvalue of the matrix $X$.


\begin{restatable}[Theorem 5.1.1~\cite{tropp2015introduction}]{relemma}{troop} \label{lemma:troop}
Let $X_1, X_2, \ldots, X_n$ be a finite sequence 
of identically distributed independent, random, 
symmetric matrices with common dimension $k$. 
Assume $0 \preceq X \preceq R\cdot I$ and $\mu_{\min}I \preceq \E{X} \preceq \mu_{\max}I$.
With probability at least $1-\delta$, 
\[ \lambda_{\min}\left(\frac{1}{n}\sum_{i=1}^n X_i\right) \ge  \mu_{\min} - \sqrt{\frac{2R\mu_{\min}\log(\frac{k}{\delta})}{n}} \,. \addeq\label{eq:eigentropp}
\]
\end{restatable}

\errone*

\begin{proof}
We present our proof in two steps. Step-1 is the non-probabilistic part, i.e., bounding the error $\norm{\wf - \w_{\f}}$ in terms of the gradient difference  $\norm{\nabla_{\w} \LL(\w_{\f}, \f) - \nabla_{\w} \LL_m(\w_{\f}, \f)}$. This step uses Taylor's expansion upto second order terms for empirical log-likelihood around the true $\w^*$. 
Step-2 involves deriving a concentration on the gradient difference using the Lipschitz property implied by Condition~\ref{cond:tau}. Combining these two steps along with Lemma~\ref{eq:troop} concludes the proof. Now we detail each of these steps.

\textbf{Step-1.} We represent the empirical Negative Log-Likelihood (NLL) function with $\LL_m$
by absorbing the negative sign to simplify notation. 
Using a Taylor expansion, we have 
$$ \LL_m(\wf, \f) = \LL_m(\w_{\f}, \f) + \inner{\nabla_{\w} \LL_m(\w_{\f}, \f)}{\wf - \w_{\f}} + \frac{1}{2} (\wf - \w_{\f})^T \nabla^2_{\w}\LL_m (\widetilde\w, \fc)(\wf - \w_{\f})\,, $$
where $\widetilde\w \in [\wf, \w_{\f}]$. 
With the assumption $\f^T \w_{\f} \ge \tau$, 
we have $\nabla^2_{\w}\LL_m (\widetilde\w, \f) \ge \frac{\tau^2}{\min \ps(y)^2} \nabla^2_{\w}\LL_m (\w_{\f}, \f)$.
Let $\kappa = \frac{\tau^2}{\min \ps(y)^2}$. Using this we get,  
$$ \LL_m(\wf, \f) \ge \LL_m(\w_{\f}, \f) + \inner{\nabla_{\w} \LL_m(\w_{\f}, \f)}{\wf - \w_{\f}} + \frac{\kappa}{2} (\wf - \w_{\f})^T \nabla^2_{\w}\LL_m (\w_{\f}, \f)(\wf - \w_{\f}) $$
$$ \underbrace{\LL_m(\wf, \f) - \LL_m(\w_{\f}, \f)}_{\RN{1}} - \inner{\nabla_{\w} \LL_m(\w_{\f}, \f)}{\wf - \w_{\f}}  \ge \frac{\kappa}{2} (\wf - \w_{\f})^T \nabla^2_{\w}\LL_m (\w_{\f}, \f)(\wf - \w_{\f}) \,, $$
where term-$\RN{1}$ is less than zero as $\wf$ is the minimizer of empirical NLL $\LL_m(\wf, \f)$.
Ignoring term-$\RN{1}$ and re-arranging a few terms we get:  
$$ - \inner{\nabla_{\w} \LL_m(\w_{\f}, \f)}{\wf - \w_{\f}}  \ge \frac{\kappa}{2} (\wf - \w_{\f})^T \nabla^2_{\w}\LL_m (\w_{\f}, \f)(\wf - \w_{\f}) \,, $$
With first order optimality on $\w_{\f}$, $\inner{\nabla_{\w} \LL(\w_{\f}, \f)}{\wf - \w_{\f}} \ge 0$. Plugging in this, we have,  
$$ \inner{\nabla_{\w} \LL(\w_{\f}, \f) - \nabla_{\w} \LL_m(\w_{\f}, \f)}{\wf - \w_{\f}}  \ge \frac{\kappa}{2} (\wf - \w_{\f})^T \nabla^2_{\w}\LL_m (\w_{\f}, \f)(\wf - \w_{\f}) \,, $$
Using Holder's inequality on the LHS we have, 
$$ \norm{\nabla_{\w} \LL(\w_{\f}, \f) - \nabla_{\w} \LL_m(\w_{\f}, \f)} \norm{\wf - \w_{\f}}  \ge \frac{\kappa}{2} (\wf - \w_{\f})^T \nabla^2_{\w}\LL_m (\w_{\f}, \f)(\wf - \w_{\f})\,.$$
Let $\shmin$ be the minimum eigenvalue of $\nabla^2_{\w}\LL_m (\w^*, \fc)$. Using the fact that $(\wf - \w_{\f})^T \nabla^2_{\w}\LL_m (\w_{\f}, \f)(\wf - \w_{\f}) \ge \widehat \sigma_{\min} \norm{\wf - \w_{\f}}^2$, we get,     
\begin{equation} \label{eq:ineq1}
\norm{\nabla_{\w} \LL(\w_{\f}, \f) - \nabla_{\w} \LL_m(\w_{\f}, \f)}  \ge \frac{\kappa \shmin}{2} \norm{\wf - \w_{\f}}  \,. 
\end{equation}


\textbf{Step-2.} The empirical gradient is $\nabla_{\w} \LL_m(\w_{\f}, \f)= \sum_{i=1}^m \frac{\nabla_{\w} \LL_1 (\x_i, \w_{\f}, \f) }{m}$
where \allowdisplaybreaks{$
\nabla \mathcal{L}_1 (\x_i, \w_{\f}, \f) =  \begin{bmatrix} 
    \frac{f_1(x_i)}{\inner{\f(\x_i)}{\w_{\f}}} 
    \dots  
    \frac{f_l(x_i)}{\inner{\f(\x_i)}{\w_{\f}}} 
    \dots 
    \frac{ f_{k}(x_i) }{\inner{\f(\x_i)}{\w_{\f}}} 
    \end{bmatrix}_{(k)}
$.}
With the lower bound $\tau$ on $\f^T\w_{\f}$, 
we can upper bound the gradient terms as
$$\norm{\nabla_{\w} \LL_1(\x, \w_{\f}, \f)} \le  \frac{\norm{\f}}{\tau} \le  \frac{\norm{\f}_1}{\tau} \le \frac{1}{\tau} \,. $$

As the gradient terms decompose and are independent, 
using Hoeffding's inequality we have with probability at least $1-\frac{\delta}{2}$, 
\begin{equation} \label{eq:ineq2}
\norm{\nabla_{\w} \LL(\w_{\f}, \f) - \nabla_{\w} \LL_m(\w_{\f}, \f)}  \le \frac{1}{2\tau}\sqrt{\frac{\log(4/\delta)}{m}}   \,.
\end{equation}

Let $\sigma_{\f, \w_{\f}}$ be the minimum eigenvalue of $\nabla^2_{\w}\LL (\w_{\f}, \f)$.
Using lemma~\ref{lemma:troop}, with probability at least $1-\frac{\delta}{2}$, 
\begin{equation} \label{eq:troop}
\frac{\shmin}{\smin} \ge 1- \tau \sqrt{\frac{\log(2k/\delta)}{m}} \,.
\end{equation}

Plugging~\eqref{eq:ineq2} and~\eqref{eq:troop} in \eqref{eq:ineq1}, 
and applying a union bound, we conclude that with probability at least $1-\delta$, 
\begin{align*}
\|\wf - \w_{\f} \|_2 &\le \frac{1}{\kappa\tau} \Big(\smin- \smin\tau \sqrt{\frac{\log(2k/\delta)}{m}}\Big)^{-1} \Big(\sqrt{\frac{\log(4/\delta)}{m}}\Big) \\
&\le \frac{1}{\kappa \tau} \frac{1}{\smin} \Big(1 + \tau \sqrt{\frac{\log(2k/\delta)}{m}}\Big) \sqrt{\frac{\log(4/\delta)}{m}} \,.
\end{align*}
Neglecting the order $m$ term and letting $c = \frac{1}{\kappa\tau} $, we have
$$\norm{\wf - \w_{\f}} \le \frac{c}{\smin} \sqrt{\frac{\log(4/\delta)}{m}} \,.$$
\end{proof}

\errtwo*

\begin{proof}
We present our proof in two steps. Note, all calculations are non-probabilistic. 
Step-1 involves bounding the error $\norm{\w_{\f} - \w^*}$ in terms of the gradient difference  $\norm{\nabla_{\w} \LL(\w^*, \fc)-\nabla_{\w} \LL(\w^*, \f)}$. This step uses Taylor's expansion on $\LL(\w_{\f}, \f)$ upto the second orderth term for population log-likelihood around the true $\w^*$. 
Step-2 involves deriving a bound on the gradient difference in terms of the difference $\norm{\f - \fc}$ using the Lipschitz property implied by Condition~\ref{cond:tau}. Further, for a crude calibration choice of $\fc(\x) =\ps (\cdot | \x)$, the gradient difference can be bounded by miscalibration error. We now detail both of these steps. 

\textbf{Step-1.} Similar to Lemma~\ref{lemma:bound-finite}, 
we represent with $\LL$ by absorbing the negative sign to simplify notation.
Using the Taylor expansion, we have
\[
\LL(\w_{\f}, \f) \ge \LL(\w^*, \f) + \inner{\nabla_{\w} \LL(\w^*, \f)}{\w_{\f} - \w^*} + \frac{1}{2} (\w_{\f} - \w^*)^T \nabla^2_{\w}\LL (\widetilde\w, \f)(\w_{\f} - \w^*) \,, 
\]
where $\widetilde\w \in [\w_{\f}, \w^*] $. 
With the assumption $\f^T\w^*\ge \tau$, we have 
$\nabla^2_{\w}\LL (\widetilde \w, \f) \ge \frac{\tau^2}{\min \ps(y)^2} \nabla^2_{\w}\LL(\w^*, \f)$ 
.  Let $\kappa = \frac{\tau^2}{\min \ps(y)^2}$. Using this we get,

$$ \LL(\w_{\f}, \f) \ge \LL(\w^*, \f) + \inner{\nabla_{\w} \LL(\w^*, \f)}{\w_{\f} - \w^*} + \frac{\kappa}{2}(\w_{\f} -\w^*)^T \nabla^2_{\w}\LL (\w^*, \f)(\w_{\f} - \w^*)$$
$$ \underbrace{\LL(\w_{\f}, \f) - \LL(\w^*, \f)}_{\RN{1}} \ge \inner{\nabla_{\w} \LL(\w_{\f}, \f)}{\w_{\f} - \w^*} + \frac{\kappa}{2}(\w_{\f} - \w^*)^T \nabla^2_{\w}\LL(\w^*, \f)(\w_{\f} - \w^*) \,,$$

where term-$\RN{1}$ is less than zero as $\w_{\f}$ is the minimizer of NLL $\LL(\w, \f)$.
Ignoring that term and re-arranging a few terms we get
$$  -\inner{\nabla_{\w} \LL(\w^*, \f)}{\w_{\f} - \w^*} \ge \frac{\kappa}{2}(\w_{\f} - \w^*)^T \nabla_{\w}^2\LL (\w^*, \f)(\w_{\f} - \w^*) \,.$$

With first order optimality on $\w^*$, $\inner{\nabla_{\w} \LL(\w^*, \fc)}{\w_{\f} - \w^*} \ge 0$.
Using this we have: 
$$  \inner{\nabla_{\w} \LL(\w^*, \fc)}{\w_{\f} - \w^*} -\inner{\nabla_{\w} \LL(\w^*, \f)}{\w_{\f} - \w^*} \ge \frac{\kappa}{2}(\w_{\f} - \w^*)^T \nabla^2_{\w}\LL (\w^*, \f)(\w_{\f} - \w^*) \,,$$
$$  \inner{\nabla_{\w} \LL(\w^*, \fc)-\nabla_{\w} \LL(\w^*, \f)}{\w_{\f} - \w^*} \ge \frac{\kappa}{2}(\w_{\f} - \w^*)^T \nabla^2_{\w}\LL (\w^*, \f)(\w_{\f} - \w^*) \,.$$

As before, let $\sfmin$ be the minimum eigenvalue of $\nabla^2_{\w}\LL (\w^*, \f)$.
Using the fact that $(\w_{\f} - \w^*)^T \nabla^2_{\w}\LL(\w^*, \f)(\w_{\f} - \w^*) \ge \sfmin \norm{\w_{\f} - \w^*}^2$, we get
$$  \inner{\nabla_{\w} \LL(\w^*, \fc) - \nabla_{\w} \LL(\w^*, \f)}{\w_{\f} - \w^*} \ge \frac{\kappa \sfmin}{2}\norm{\w_{\f} - \w^*}^2 \,.$$

Using Holder's inequality on the LHS and re-arranging terms gives
\begin{equation} \label{eq:ineq3}
\norm{\nabla_{\w} \LL(\w^*, \fc)-\nabla_{\w} \LL(\w^*, \f)} \ge \frac{\kappa \sfmin }{2} \norm{\w_{\f} -\w^*}  \,.
\end{equation}

\textbf{Step-2.} By lower bound assumptions $\fc^T \w^* \ge \tau$ and $\f^T \w^* \ge \tau$, we have
\[ \norm{\nabla_{\w} \LL(\w^*, \fc) - \nabla \LL( \w^*, \f)} \le \EE{t}{\norm{\nabla \LL_1(\x, \w^*, \fc) - \nabla \LL_1( \x, \w^*, \f)}} 
\le \frac{1}{\tau^2} \EE{t}{ \norm{ \fc(\x) - \f(\x) } } \,, \numberthis \label{eq:err2-bound}
\]
where the first inequality is implied by Jensen's inequality
and the second is implied by the Lipschitz property of the gradient. 
Further, we have 
\begin{align*}
  \mathbb{E}_{t}\left[\norm{ \fc (\x) - \f(\x) }\right] 
  &= \mathbb{E}_{s}\left[ \frac{\pt(\x)}{\ps(\x)} \norm{ \fc (\x) - \f(\x) }\right]  \\ 
  &\le \mathbb{E}_{s}\left[ \max_y \frac{\pt(y)}{\ps(y)} \norm{ \fc (\x) - \f(\x) }\right] \\
  &\le \max_y \frac{\pt(y)}{\ps(y)} \mathbb{E}_{s}\left[ \norm{ \fc (\x) - \f(\x) }\right] \,. \numberthis \label{eq:term2_bound}
\end{align*}

Combining equations \eqref{eq:ineq3}, \eqref{eq:err2-bound}, and \eqref{eq:term2_bound}, we have

\[ 
 \norm{\w_{\f} - \w^*} \le  \frac{2}{\kappa \sfmin \tau^2} \max_y \frac{\pt(y)}{\ps(y)} \mathbb{E}_{s}\left[ \norm{ \fc (\x) - \f(\x) }\right] \,. \numberthis \label{eq:err2_bound1}
\]

Further, if we set $\f_c(\x) = p_s(\cdot | \f(\x))$, 
which is a calibrated predictor according to Proposition~\ref{prop:ltt-calib}, 
we can bound the error on the RHS in terms of the calibration error of $\f$. Moreover, in the label shift estimation problem, we have the assumption that $p_s(y)\ge c >0$ for all $y$. Hence, we have $\max_y p_t(y)/p_s(y) \le 1/c$.    
Using Jensen's inequality, we get 
\[
\mathbb{E}_{s}{\norm{ \fc(\x) - \f(\x) }} \le \left(\mathbb{E}_{s}{\norm{ \fc(\x) - \f(\x) }^2}\right)^{\frac{1}{2}} = \CE{\f} \,. \numberthis \label{eq:calib-err2}
\] 

Plugging \eqref{eq:calib-err2} back in \eqref{eq:err2_bound1},we get the required upper bound. 
\end{proof}

\begin{manualtheorem}{3}
For any $\w \in \W$, we have 
$
\sigma_{\f,\w} \ge p_{s, \min} \sigma_{\f}
$
where $\sigma_{\f}$ is the minimum eigenvalue of $\EE{t}{\f(\x)\f(\x)^T}$ 
and $p_{s, \min}=\min_{y\in\out} p_s(y)$. 
Furthermore, if $\f$ satisfies Condition~\ref{cond:tau}, we have
 $p_{s, \min}^2 \cdot \sigma_{\f} \le \sigma_{\f,\w} \le \tau^{-2} \cdot \sigma_{\f} \,,$
for $\w\in \{\w_{\f}, \w^*\}$.
\label{prop:eigen}
\end{manualtheorem}

\begin{proof}
For any $\vv \in \Real^k$, we have 
\begin{align*}
\vv^T \left( - \nabla^2_{\w} \LL(\w, \f) \right) \vv = 
\EE{t}{\frac{\left(\vv^T\f(\x)\right)^2}{\left(\f(\x)^T \w\right)^2}} \in \left[\frac{1}{a^2}, \frac{1}{b^2}\right] \cdot \vv^T \EE{t}{\f(\x)\f(\x)^T} \vv \,,
\end{align*}
where
\begin{align*}
a = \max_{\x:p_s(\x)>0} \f(\x)^T \w \le \frac{1}{p_{s, \min}} \,
\end{align*}
and
\begin{align*}
b = \min_{\x:p_s(\x)>0} \f(\x)^T \w \ge \tau \,
\end{align*}
if $\f$ satisfies Condition~\ref{cond:tau} and $\w \in \{\w_{\f}, \w^*\}$. Therefore,
we have 
\begin{align*}
 p_{s, \min}^2 \cdot \sigma_{\f} \le \sigma_{\f,\w} \le \tau^{-2} \cdot \sigma_{\f}
\end{align*}
for $\w\in \{\w_{\f}, \w^*\}$.

\end{proof}

\ce*
\begin{proof}
Assume regularity conditions on the model class $\G_{\theta}$ 
(injectivity, Lipschitz-continuity, twice differentiability, non-singular Hessian, and consistency) 
as in Theorem 5.23 of \citet{stein1981estimation} hold true. 
Using the injectivity property of the model class as in~\citet{kumar2019verified},
we have for all $g_1, g_2\in \G$, 
\begin{equation} \label{eq:MSE}
\MSE{g_1} - \MSE{g_2} = \CE{g_1}^2 - \CE{g_2}^2  \,.  
\end{equation}
Let $\widehat g, g^* \in \G$ be models parameterized by $\widehat \theta$ and $\theta^*$, respectively.
Using the strong convexity of the empirical mean squared error we have,
$$\nMSE{\widehat g} \ge \nMSE {g^*} + \inner{\nabla_{\theta} \nMSE{g^*}}{\widehat \theta - \theta^*} 
+ \frac{\mu^2}{2} \norm{\widehat \theta - \theta^*}^2_2 \,, $$ 
where $\mu$ is the parameter constant for strong convexity.
Re-arranging a few terms, we have 
$$ \underbrace{ \nMSE{\widehat g} - \nMSE {g^*} }_{\RN{1}} - \inner{\nabla_{\theta} \nMSE{g^*}}{\widehat \theta - \theta^*} \ge  \frac{\mu^2}{2} \|\widehat \theta - \theta^*\|^2_2   \,, $$ 
where term-$\RN{1}$ is less than zero because $\widehat \g$ 
is the empirical minimizer of the mean-squared error. 
Ignoring term-$\RN{1}$, we get: 
$$ \frac{\mu^2}{2} \|\widehat \theta - \theta^*\|^2_2  \le  - \inner{\nabla_{\theta} \nMSE{g^*}}{\widehat \theta - \theta^*} \le \norm{ \nabla_{\theta} \nMSE{g^*}} \norm{{\widehat \theta - \theta^*}} \,. $$ 

As the assumed model class is Lipschitz w.r.t. $\theta$,
the gradient is bounded by Lipschitz constant $L = c_1$. 
$\E{\nabla_{\theta} \nMSE{g^*}} = 0$ as $g^*$ is the population minimizer.
Using Hoeffding's bound for bounded functions,
we have with probability at least $1-\delta$,
\begin{equation} \label{eq:HoeffMSE} 
\|{\widehat \theta - \theta^*}\|_2 \le \frac{c_1}{\mu^2} \sqrt{\frac{\log(2/\delta)}{n}} \,.
\end{equation}
Using the smoothness of the $\MSE{g}$, we have 
\begin{equation} \label{eq:smoothness}
\MSE{\widehat g} - \MSE{g^*} \le c_2 \|\widehat \theta - \theta^*\|^2_2  \,,  
\end{equation}
where $c_2$ is the operator norm of the $\nabla^2\MSE{g^*} $.
Combining~\eqref{eq:MSE},~\eqref{eq:HoeffMSE}, and~\eqref{eq:smoothness}, 
we have for some universal constant $c = \frac{c_1c_2}{\mu^2}$ with probability at least $1-\delta$, 
\[
\CE{\widehat g}^2 - \CE{g^*}^2 \le   c \frac{\log(2/\delta)}{n} \,.
\]
\end{proof}

Moreover, with Lemma~\ref{lemma:bound-calib}, depending on the degree of the miscalibration 
and the method involved to calibrate, we can bound the $\CE{\f}$. 
For example, if using vector scaling on a held out training data for calibration,
we can use Lemma~\ref{lemma:ce} to bound the calibration error $\CE{\f}$, i.e., 
with probability at least $1-\delta$, we have
\begin{equation} \label{eq:calib_err}
\CE{\f} \le \sqrt{\min_{\g \in \G} \CE{\g \circ \f }^2 + c \frac{\log(2/\delta)}{n}} \le  \min_{\g \in \G} \CE{\g \circ \f } + \sqrt{c\frac{\log(2/\delta)}{n}}    \,.
\end{equation}
 

Plugging \eqref{eq:calib-err2} and \eqref{eq:calib_err} into \eqref{eq:err2_bound1}, 
we have with probability at least $1-\delta$ that 

$$ \norm{\w_{\f} - \w^* } \le \frac{1}{\kappa \sfmin \tau^2}\left(\|\w^*\|_2 \left( \sqrt{c\frac{\log(2/\delta)}{n}} + \min_{\g \in \G} \CE{
\g \circ \f } \right) \right)\,.  $$ 


\end{document}